\documentclass[11pt]{article}

\usepackage[preprint]{acl}

\usepackage{times}
\usepackage{latexsym}
\usepackage[T1]{fontenc}
\usepackage[utf8]{inputenc}
\usepackage{microtype}
\usepackage{inconsolata}

\usepackage{graphicx}
\usepackage{booktabs}
\usepackage{caption}     

\usepackage[accsupp]{axessibility}  

\usepackage{booktabs}   
\usepackage{multirow}   
\usepackage{amsmath}
\usepackage{mathtools}
\usepackage{makecell} 
\usepackage{enumitem}
\usepackage{graphicx}   
\usepackage{balance}
\usepackage{afterpage}
\usepackage[table]{xcolor} 
\usepackage{tikz}
\usepackage{authblk}
\usepackage{tabularx} 
\usepackage[bottom]{footmisc}

\definecolor{mindvgoldA}{HTML}{8A6225}
\definecolor{mindvgoldB}{HTML}{7A530C}
\definecolor{mindvgoldC}{HTML}{8A6225}
\definecolor{mindvgoldD}{HTML}{9A7130}
\definecolor{mindvgoldE}{HTML}{AA813C}
\definecolor{mindvgoldF}{HTML}{C0954B}
\definecolor{mindvgoldShadow}{HTML}{4A3212}
\newcommand{\MindVLetters}{%
  \textbf{%
    \textcolor{mindvgoldA}{M}%
    \textcolor{mindvgoldB}{I}%
    \textcolor{mindvgoldC}{N}%
    \textcolor{mindvgoldD}{D}%
    \textcolor{mindvgoldE}{-}%
    \textcolor{mindvgoldF}{V}%
  }%
}
\DeclareRobustCommand{\MindV}{%
  \texorpdfstring{%
    \tikz[baseline=(mindvtext.base)]{%
      \node[inner sep=0pt, outer sep=0pt, text=mindvgoldShadow, xshift=0.22pt, yshift=-0.22pt]
        {\textbf{MIND-V}};%
      \node[inner sep=0pt, outer sep=0pt] (mindvtext) {\MindVLetters};%
    }%
  }{MIND-V}%
}

\definecolor{bestcolor}{HTML}{C8E6C9}     
\definecolor{secondbestcolor}{HTML}{E8F5E9} 
\definecolor{thirdbestcolor}{HTML}{F1F8E9}  

\newcommand{\best}[1]{\colorbox{bestcolor}{\textbf{#1}}}

\newcommand{\thirdbest}[1]{\colorbox{thirdbestcolor}{#1}}

\newcommand{\runhead}[1]{\par\smallskip\noindent\textbf{#1}\quad}
\newcommand{\blockhead}[1]{\par\smallskip\noindent\textbf{#1}\par\nobreak\vspace{0.15em}\noindent}

\begin{document}

\title{\MindV{}: Hierarchical World Model for Long-Horizon Robotic Manipulation with RL-based Physical Alignment} 


\author{
  Ruicheng Zhang\textsuperscript{1} \quad
  Mingyang Zhang\textsuperscript{1} \quad
  Jun Zhou\textsuperscript{1} \quad
  Xiaofan Liu\textsuperscript{3} \\
  Zunnan Xu\textsuperscript{1,2}\thanks{Project Lead.} \quad
  Zhizhou Zhong\textsuperscript{4} \quad
  Puxin Yan\textsuperscript{4} \quad
  Haocheng Luo\textsuperscript{1} \quad
  Xiu Li\textsuperscript{1}\thanks{Corresponding author.}
  \vspace{0.5em} \\ 
   \textsuperscript{1}Tsinghua University \quad
  \textsuperscript{2}X Square Robot \quad
  \textsuperscript{3}Sun Yat-sen University \quad
  \textsuperscript{4}HKUST
}


\maketitle

\vspace{-1em}
\begin{abstract}
Scalable embodied intelligence is constrained by the scarcity of diverse, long-horizon robotic manipulation data. Existing video world models are limited to short clips of simple actions and frequently rely on manually defined trajectories. We introduce \MindV{}, a cognitive hierarchical world model for synthesizing physically plausible, logically coherent long-horizon manipulation videos. Inspired by cognitive science, \MindV{} bridges high-level reasoning with pixel-level synthesis via three components: a Semantic Reasoning Hub (SRH) that leverages a pretrained vision-language model for task planning; a Behavioral Semantic Bridge (BSB) that converts abstract instructions into domain-invariant representations; and a Motor Video Generator (MVG) for conditional video rendering. To enforce physical plausibility, we introduce a GRPO-based RL post-training stage guided by a Physical Foresight Coherence (PFC) reward, which employs V-JEPA2 as a physics referee to penalize implausible dynamics in latent space. Experiments demonstrate state-of-the-art performance in long-horizon simulation and significant utility for downstream policy learning, establishing a scalable, fully autonomous framework for embodied data synthesis.
\end{abstract}

\vspace{-0.5em}
\section{Introduction}
\label{sec:intro}
\vspace{-0.5em}
Scalable robot learning within Embodied AI~\cite{pi0} is critically bottlenecked by the scarcity of high-quality, diverse, and long-horizon robotic manipulation data~\cite{robomaster}. Video world models~\cite{IRASim,wow,cosmos} (VWM) that can simulate physical interactions and predict future outcomes offer a promising solution by potentially synthesizing an infinite stream of robotic operation videos.

\begin{figure}[t]
  \centering
  \setlength{\abovecaptionskip}{-0.02em}   
   \includegraphics[width=\linewidth]{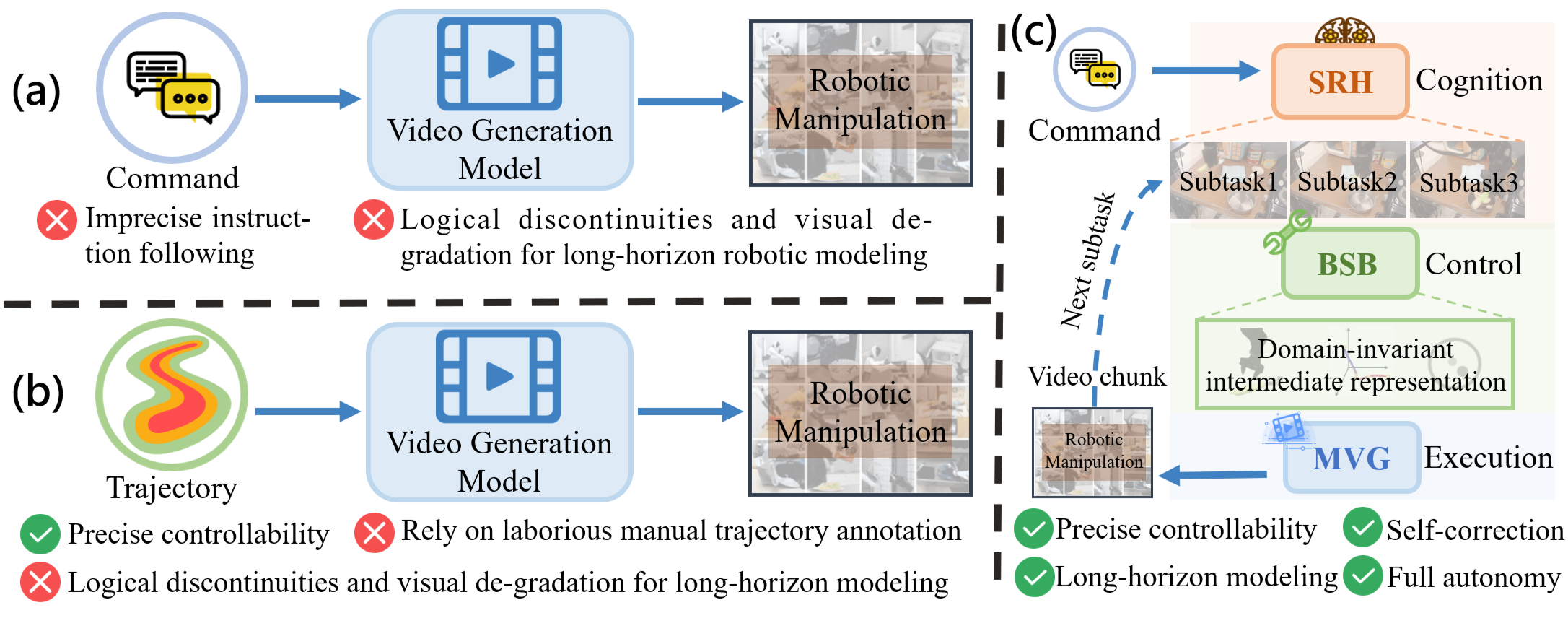}
   \caption{\textbf{Architectural comparison} of \MindV{} (c) against existing paradigms for long-horizon robotic manipulation modeling.}
   \vspace{-2em}
   \label{fig:f0}
\end{figure}

However, generating high-quality, long-horizon robotic manipulation videos that faithfully follow commands poses three core challenges:
\textbf{(1) Long-Horizon Coherence:} Maintaining causal consistency and logical flow across interconnected subtasks, where a single error can compromise the entire operation~\cite{robobrain}.
\textbf{(2) Semantic-to-Pixel Generation:} Accurately translating abstract semantic commands into spatiotemporal pixel-level interactions, placing stringent demands on semantic understanding and instruction-following fidelity~\cite{controllable}.
\textbf{(3) Physical Plausibility:} Ensuring adherence to fundamental physical laws governing collision dynamics, object permanence, and interaction forces~\cite{physical_priors}.
Existing methods address these challenges only partially. Directly fine-tuning video foundation models~\cite{hunyuanvideo} for long-horizon robotic modeling suffers from imprecise instruction following and visual degradation, as these models struggle to bridge the gap from abstract commands to pixel-level execution (Fig.~\ref{fig:f0}(a)). Trajectory-controlled generative models~\cite{draganything}, while offering enhanced controllability, sacrifice the autonomy and scalability required for large-scale data generation, functioning more as renderers than as intelligent world simulators (Fig.~\ref{fig:f0}(b)).


Inspired by the hierarchical theory of human motor control~\cite{Neurobiological}, we draw an analogy to how the brain executes complex tasks: high-level cognitive centers such as the cerebral cortex handle intent understanding and abstract planning, while low-level motor systems such as the cerebellum translate these plans into precise motor commands, with specialized neural pathways bridging the two~\cite{Hierarchical}.

Embracing this paradigm, we introduce \MindV{}, a cognitive hierarchical world model for simulating physically plausible, logically coherent long-horizon robotic manipulation. \MindV{} is built upon three core components (Fig.~\ref{fig:f0}(c)): (1) a Semantic Reasoning Hub (SRH) that performs high-level task planning via a pretrained Vision-Language Model (VLM); (2) a Behavioral Semantic Bridge (BSB) that translates abstract plans into structured, domain-invariant representations; and (3) a Motor Video Generator (MVG) that synthesizes physically realistic manipulation videos conditioned on the BSB. By decomposing cognition and execution into explicit hierarchical stages, \MindV{} effectively bridges high-level reasoning with pixel-level synthesis.

To align the MVG with physical laws, we introduce a GRPO-based~\cite{flow_grpo,zhang2026kvpo} post-training stage~\cite{marl-snake} guided by a Physical Foresight Coherence (PFC) reward, which employs a pretrained world model as a physics referee to quantify the dynamic coherence of generated videos in latent space.
We further propose Staged Visual Future Rollouts, a test-time optimization strategy that decomposes long-horizon planning into locally optimal decisions. At each subtask transition, \MindV{} simulates multiple candidate futures and selects the most coherent continuation via a propose-verify-refine loop, suppressing error propagation and semantic drift.

Our main contributions are as follows:
\begin{itemize}
    \item \MindV{} is, to the best of our knowledge, the first hierarchical video world model for long-horizon robotic manipulation, bridging high-level task planning and pixel-level synthesis through a three-tier architecture comprising the SRH, BSB, and MVG.
    \item We present Staged Visual Future Rollouts, a test-time optimization strategy that decomposes long-horizon generation into locally optimal decisions via a propose-verify-refine loop at each subtask transition, mitigating error accumulation and improving generation robustness.
    \item We propose a GRPO post-training stage guided by a Physical Foresight Coherence (PFC) reward, which employs a pretrained world model as a physics referee to score generated dynamics in latent space, steering the MVG toward physically plausible outputs.
    \item Extensive experiments demonstrate state-of-the-art performance in long-horizon manipulation video synthesis and significant utility for downstream policy learning, establishing a scalable and autonomous paradigm for robotic world modeling.
\end{itemize}

\begin{figure*}[t]
  \centering
  \setlength{\abovecaptionskip}{-0.02em}   
   \includegraphics[width=0.95\textwidth]{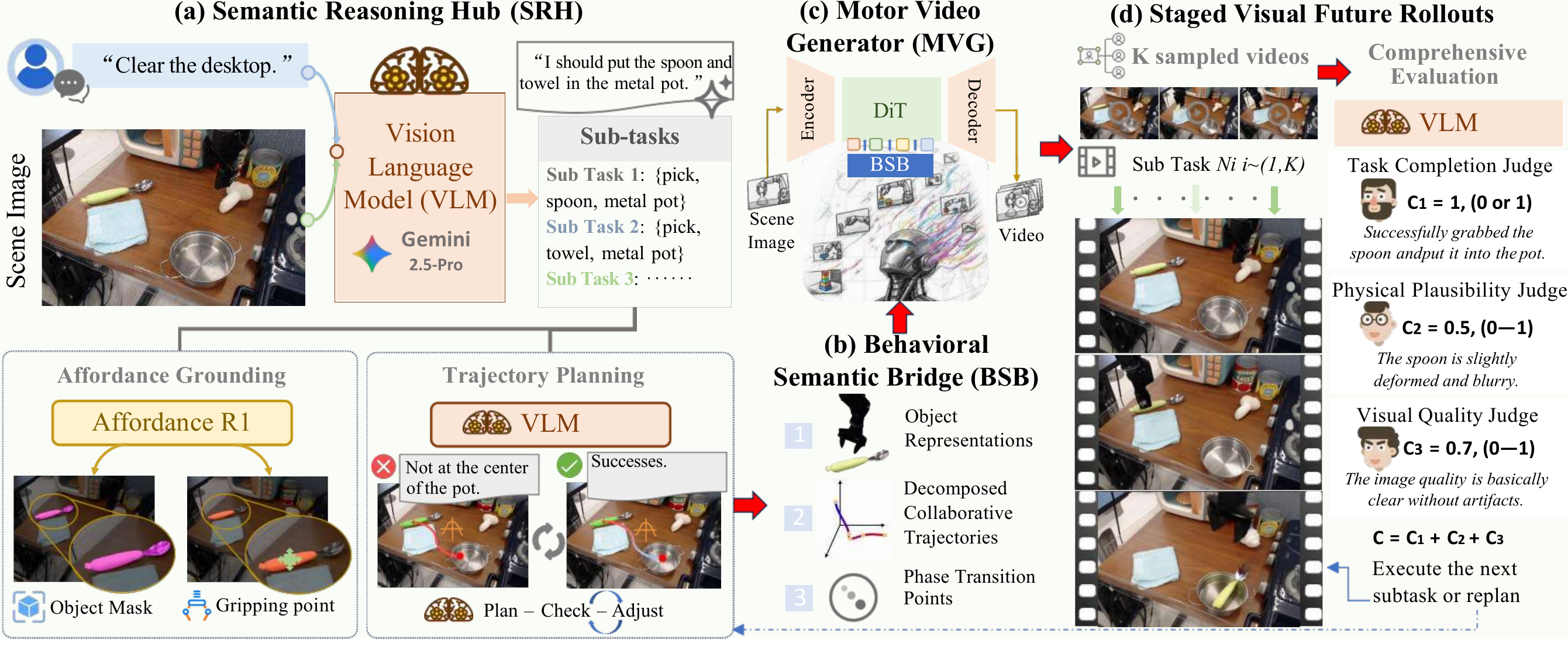}

\caption{\textbf{Overview of \MindV{}: a cognitive hierarchical world model for long-horizon robotic manipulation.} }
   \label{fig:f1}
   \vspace{-1.0em}
\end{figure*}

\vspace{-1em}
\section{Related Work}
\vspace{-0.5em}
\subsection{Video World Models for Embodied AI}
\vspace{-0.25em}
Scalable embodied intelligence critically depends on large-scale data, yet collecting real-world robotic demonstrations is costly and labor-intensive. Video World Models (VWMs)~\cite{zhang2026robostereo}, which predict future states from current observations, have emerged as an efficient alternative for synthesizing photorealistic training data. UniPi~\cite{UniPi} and AVDC~\cite{AVDC} frame robotic planning as a text-to-video generation problem, while WoW~\cite{wow}, DreamDojo~\cite{dreamdojo}, and RoboDreamer~\cite{robodreamer} learn latent physical dynamics from large-scale video interaction data to achieve compositional generalization. Despite strong semantic understanding, these methods lack fine-grained control over manipulation execution~\cite{controllable}, which can cause logical failures and physical inconsistencies in long-horizon settings. A complementary line of work, including IRASim~\cite{IRASim}, Cosmos~\cite{cosmos}, and RoboMaster~\cite{robomaster}, employs explicit trajectory guidance for more precise action modeling, but requires dense manual annotations such as motion paths and object masks, limiting scalability and autonomy. \MindV{} addresses both limitations through a hierarchical architecture that autonomously decomposes high-level commands into executable generator instructions, enabling long-horizon, high-fidelity manipulation video synthesis without additional manual supervision.

\vspace{-0.5em}
\subsection{Controllable Video Generation}
\vspace{-0.25em}
Recent advances in diffusion-based generation have intensified interest in controllable methods that faithfully translate user intent into visual content. Existing approaches span a spectrum of conditioning modalities, ranging from high-level semantic signals such as text prompts~\cite{wan22} to low-level structural inputs including masks~\cite{draganything}, trajectories~\cite{zo3t}, sketches~\cite{ma2024followyouremoji}, and pose estimates~\cite{hunyuanportrait}. This spectrum reflects a fundamental trade-off: high-level conditions offer intuitive guidance but are prone to semantic drift and fidelity degradation in long-horizon, multi-stage tasks, whereas low-level conditions afford precise spatiotemporal control but impose a heavy annotation burden. \MindV{} resolves this dilemma through a hierarchical VWM framework that bridges high-level reasoning and pixel-level synthesis. The Semantic Reasoning Hub (SRH) interprets abstract user intent, while the Behavioral Semantic Bridge (BSB) automatically converts it into structured, multi-dimensional representations for the generator, achieving high semantic fidelity and precise control in long-horizon world modeling without auxiliary manual annotations.


\vspace{-0.5em}
\section{Method}
\vspace{-0.5em}

\subsection{Framework}
\label{sec:framework}
\vspace{-0.25em}
As illustrated in Fig.~\ref{fig:f1}, \MindV{} implements a top-down pipeline from high-level cognition to concrete visual representation. First, the \textit{Semantic Reasoning Hub (SRH)} decomposes a long-horizon task into atomic subtasks based on initial observations and user instructions. For each subtask, the SRH then employs vision modules for affordance localization and trajectory planning to construct a structured, domain-invariant intermediate representation, termed the \textit{Behavioral Semantic Bridge (BSB)}. This representation guides the \textit{Motor Video Generator (MVG)} in synthesizing a photorealistic video sequence. A closed-loop feedback mechanism via \textit{Staged Visual Future Rollouts} returns the generated results to the SRH for evaluation and potential re-planning. A detailed description of each component follows.

\begin{figure*}[t]
  \centering
  \setlength{\abovecaptionskip}{-0.02em}   
   \includegraphics[width=0.88\textwidth]{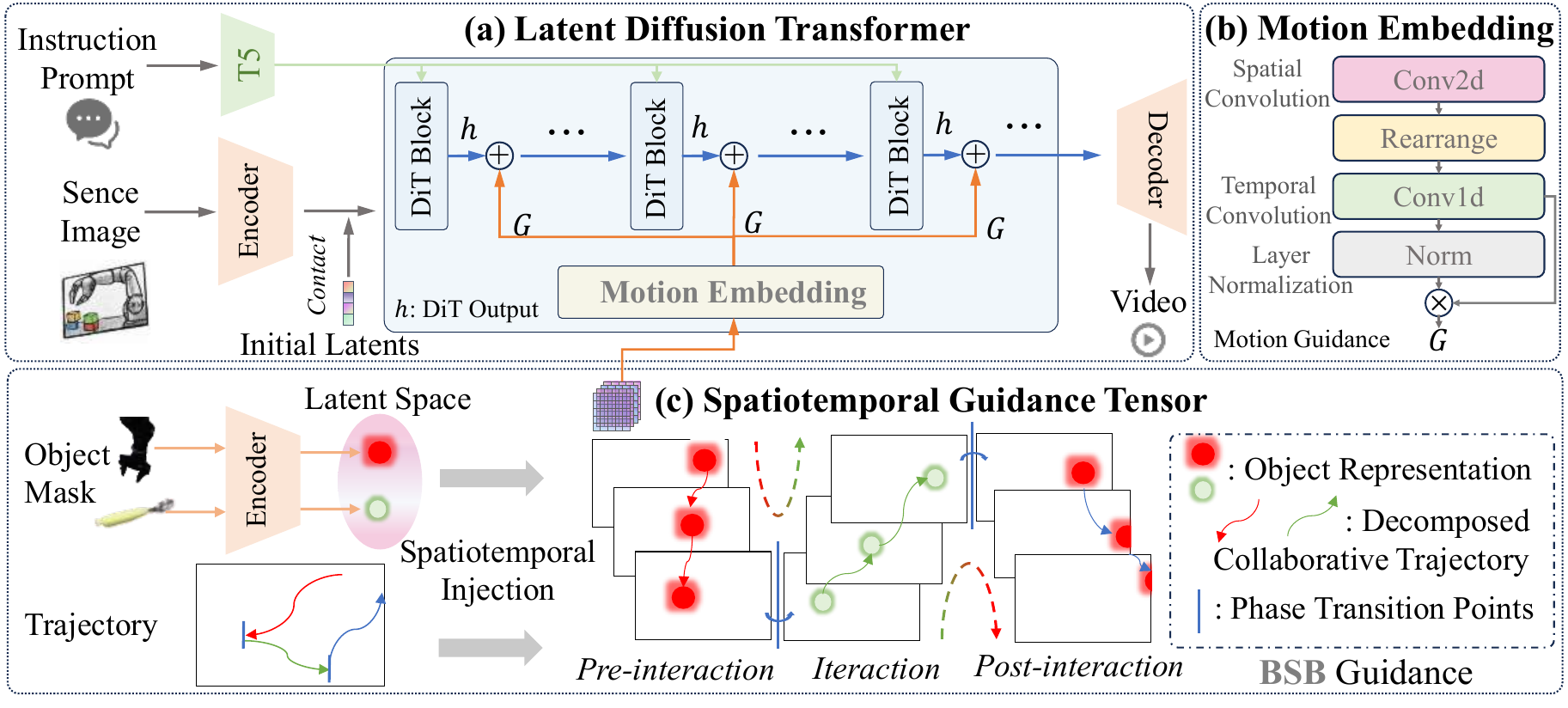}

   \caption{\textbf{Architecture of the Motor Video Generator (MVG)}. Serving as the simulation engine, the MVG translates abstract intents (BSB) into concrete visual futures. The process initiates with encoding the BSB's semantic representation into the (c) Spatiotemporal Guidance Tensor, which embeds the visual features of the active agent along its planned trajectory. This guidance is then processed by the (b) Motion Embedding module to produce refined motion signals ($G$). These signal are finally injected into the (a) Latent Diffusion Transformer, conditioning the denoising process to ensure the synthesized video strictly adheres to the intended physical dynamics.
}
   \label{fig:f2}
   \vspace{-0.5em}
\end{figure*}

\blockhead{Semantic Reasoning Hub (SRH)}
As the cognitive core of \MindV{}, the SRH translates abstract semantics into actionable geometric signals by synergizing two components: a pretrained Vision-Language Model (VLM, e.g., Gemini-2.5-Pro~\cite{gemini25}) for long-horizon planning and semantic reasoning, and an affordance-based visual localizer (e.g., Affordance-R1~\cite{affordance-r1}) for grounding plans with physical common sense.

Given an initial observation $I_0$ and a long-horizon instruction $L$ (e.g., ``clean the desktop''), the VLM analyzes the scene and decomposes $L$ into an ordered sequence of atomic subtasks, each defined by a tuple $\text{SubTask}_i = \{\text{ActionType}_i, \text{Object}_i, \text{Destination}_i\}$ specifying the action primitive, manipulation target, and goal location. This structured decomposition provides a symbolic foundation for precise downstream control. For each subtask, the affordance localizer identifies the object segmentation mask $M_{\text{obj}}$ and predicts functional interaction points $P_{\text{obj}}$ (e.g., the handle of a cup). Based on this affordance information, the VLM plans a physically plausible trajectory, parameterized as a smooth curve and discretized into frame-aligned waypoints.


\blockhead{Behavioral Semantic Bridge (BSB)}
The BSB is a structured, domain-invariant intermediate representation that translates symbolic outputs from the SRH into an actionable conditioning format for the MVG. It comprises three elements:

\begin{itemize}[leftmargin=*,topsep=0.25em,itemsep=0.25em,parsep=0pt]
\vspace{-0.25em}
    \item \textbf{Object Representation:} Segmentation masks for the manipulated object ($M_{\text{obj}}$) and robot arm ($M_{\text{rob}}$), encoded via a VAE and injected at designated spatiotemporal locations during generation to maintain consistent object identity.
\vspace{-0.25em}
    \item \textbf{Decomposed Collaborative Trajectory:} The trajectory is partitioned into three phases: pre-interaction ($T_{\text{pre}}$, arm approach), interaction ($T_{\text{interact}}$, object manipulation), and post-interaction ($T_{\text{post}}$, arm retraction), clearly defining the active agent and objective at each stage.
    \item \textbf{Phase Transition Points:} A triplet of frame indices $(F_{\text{pre}}, F_{\text{interact}}, F_{\text{post}})$ allocating a specific duration to each phase, ensuring natural motion dynamics and appropriate emphasis on the core physical interaction.
\vspace{-0.5em}
\end{itemize}

By decoupling task logic from visual appearance, the BSB achieves domain invariance, enhancing generalization to novel environments and tasks.

\blockhead{Motor Video Generator (MVG)}
As depicted in Fig.~\ref{fig:f2}, the MVG functions as a learnable physics engine. Built upon a Diffusion Transformer (DiT) backbone~\cite{DiT}, it is trained to synthesize high-fidelity video sequences that strictly adhere to the kinematic constraints defined by the BSB. To enforce this control, the MVG first encodes the BSB's object representation into a spatiotemporal guidance tensor of size ($T \times C \times H \times W$). This tensor dynamically embeds the visual features of the active agent (arm or manipulated object) onto its planned path across the time dimension. A motion embedding module integrates this guidance into the DiT backbone during denoising process. The module employs spatiotemporal convolutions to encode the guidance tensor into a feature representation ${G}$. Within each DiT block, this representation is fused with the video's intermediate hidden state $h$ via additive fusion:
\begin{equation}
    h_{\text{new}} = h + \text{norm}(G) \cdot G,
\end{equation}
where $\text{norm}(\cdot)$ denotes Group Normalization to stabilize training. This continuous injection of kinematic constraints compels the model to adhere to the specified trajectory throughout the denoising process, yielding a final video that is both spatiotemporally precise and visually coherent.

\vspace{-0.5em}
\subsection{Staged Visual Future Rollouts}
\label{tto}
\vspace{-0.25em}
Long-horizon video generation is susceptible to error accumulation, where minor deviations in early subtasks cascade into overall task failure~\cite{robridge}. To mitigate this, we reframe inference as a search process over the world model's latent space, introducing \textit{Staged Visual Future Rollouts}: a test-time optimization strategy that emulates System-2 thinking~\cite{system2} by deliberating over candidate futures at critical decision points. As illustrated in Fig.~\ref{fig:f1}(d), the strategy operates as a propose-verify-refine loop at each subtask transition. Rather than committing to a single plan, the SRH proposes $K$ semantically diverse candidate plans (i.e., BSB variants). The MVG rolls out each plan in parallel into short-horizon video clips $\{V_1, \dots, V_K\}$. The VLM then acts as a critic, evaluating each rollout on task success, physical plausibility, and visual quality.


If the top-scoring rollout $V_{\text{top}}$ meets a predefined success threshold, it is selected and the process advances to the next subtask. Otherwise, the VLM provides structured textual feedback identifying the failure modes (e.g., ``end position error''), which instructs the SRH to re-plan and propose a refined set of masks and trajectories. This iterative propose-verify-refine cycle transforms \MindV{} from a feed-forward predictor into a self-correcting agent, improving robustness and success rate in long-horizon task simulation.

\begin{figure}[t]
  \centering
  \setlength{\abovecaptionskip}{-0.02em}   
   \includegraphics[width=\linewidth]{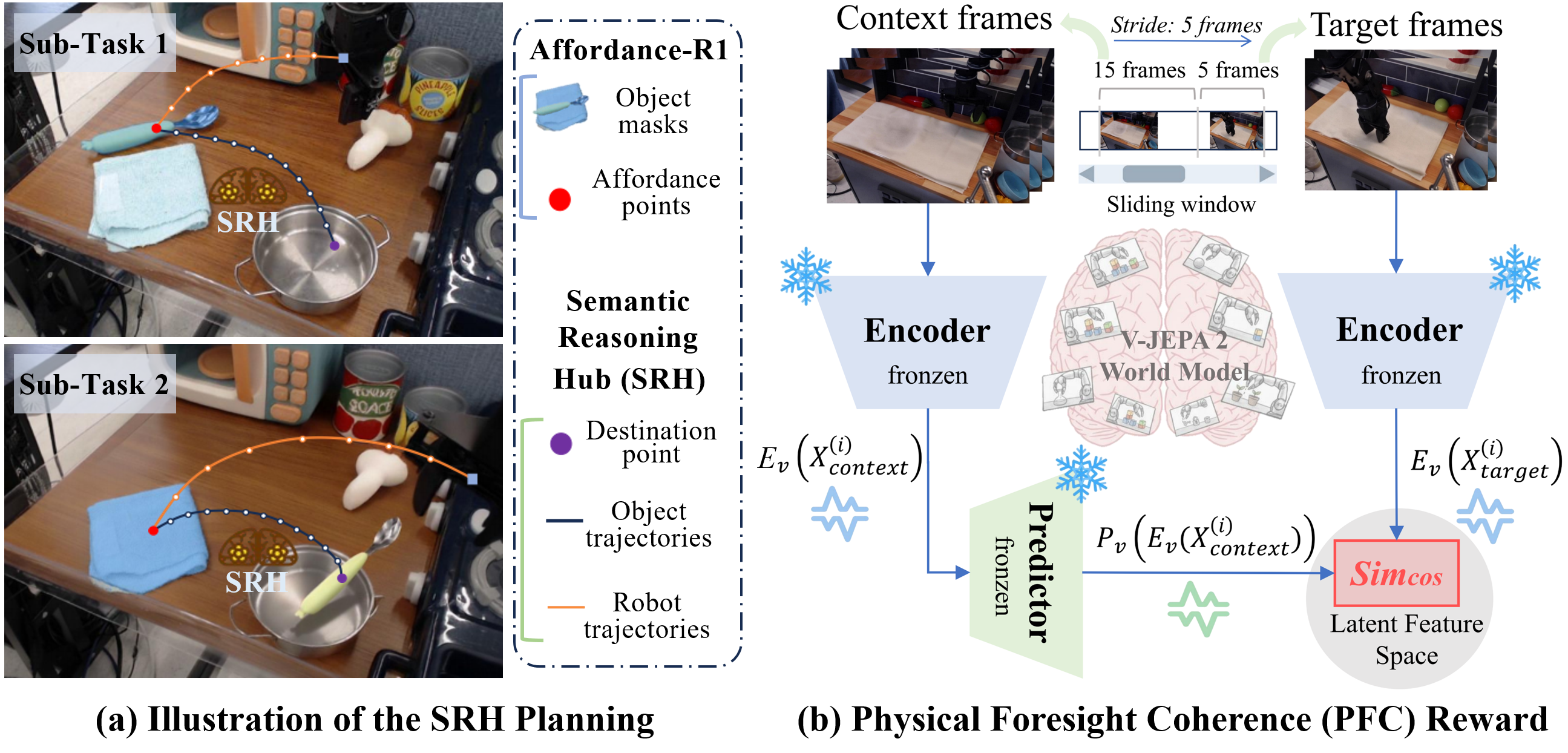}

   \caption{\textbf{(a) Illustration of the SRH Planning.} \textbf{(b) Physical Foresight Coherence (PFC) Reward.} The PFC leverages a frozen V-JEPA2 world model as a “Physics Referee” to predict the latent evolution of future states. The reward quantifies the alignment between the generated video's dynamics and V-JEPA2's internal physical laws, computed via cosine similarity in the predictive latent space.}
   \label{fig:f3}
   \vspace{-1.5em}
\end{figure}

\vspace{-0.5em}
\subsection{MVG Training: From SFT to RL}
\vspace{-0.25em}
To establish the MVG as a physically grounded simulator, we adopt a two-stage training paradigm: Supervised Fine-Tuning (SFT) to adapt a pretrained video model to the robotics domain, followed by a GRPO post-training that aligns the model's dynamics with physical laws and aesthetic standards.

\blockhead{Stage 1: Supervised Fine-Tuning (SFT)}
SFT trains the MVG to learn the mapping from BSB representations to coherent pixel-level execution, minimizing the standard BSB-conditioned denoising objective:
\begin{equation}
\mathcal{L}_{\text{SFT}}(\theta) = \mathbb{E}_{\epsilon \sim \mathcal{N}(0, I),\, t} \left[ \| \epsilon - \epsilon_\theta(x_t, t, \text{BSB}) \|^2 \right],
\end{equation}
where $x_t$ is the noised video at timestep $t$. This stage yields a reference policy $\pi_{\text{ref}}$ for subsequent alignment. Notably, training on short subtask videos suffices, as the hierarchical framework enables generalization to arbitrary long-horizon tasks.

\blockhead{Stage 2: Physical Alignment via GRPO Post-Training}
\label{sec:GRPO}
SFT alone cannot guarantee physical plausibility or aesthetic quality, objectives ill-suited to conventional loss functions~\cite{flow_grpo}. We therefore introduce an RL alignment stage that models the denoising process as a Markov Decision Process and employs Group Relative Policy Optimization (GRPO)~\cite{flow_grpo}. Optimization is guided by a composite reward:
\begin{equation}
R(x_0) = w_p \cdot R_{\text{physics}}(x_0) + w_a \cdot R_{\text{aesthetic}}(x_0),
\label{eq:reward}
\end{equation}
with $w_p = 0.2$ and $w_a = 1$; ablation details are provided in Sec.~\ref{ablation}.

\runhead{Physical Foresight Coherence (PFC) Reward ($R_{\text{physics}}$)}
We leverage V-JEPA2~\cite{vjepa2}, pretrained via self-supervision on large-scale video data and fine-tuned on robotics datasets, as a physics referee. Its learned world dynamics enable accurate latent-space prediction of future states. For each generated video, a sliding window computes the local physical plausibility score $s_i$ as the cosine similarity between V-JEPA2's latent prediction and the actual future (Fig.~\ref{fig:f3}):
\begin{equation}
s_i = \text{sim}_{\cos}\!\left( P_v(E_v(x_{\text{context}}^{(i)})),\, E_v(x_{\text{target}}^{(i)}) \right),
\end{equation}
where $E_v$ and $P_v$ denote V-JEPA2's visual encoder and predictor, respectively. To concentrate optimization on the most severe physical violations, a softmax-weighted aggregation assigns higher weights to windows with larger errors $(1 - s_i)$:

\vspace{-1.25em}
\begin{small}
\begin{equation}
R_{\text{physics}}(x_0) = \sum_{i=1}^{N_w} \frac{\exp((1-s_i)/\tau)}{\sum_{j=1}^{N_w} \exp((1-s_j)/\tau)} \cdot s_i,
\end{equation}
\end{small}
\vspace{-1em}

where temperature $\tau$ controls focus sharpness: lower values concentrate the reward on the worst-offending window. The PFC reward thus reframes physical evaluation as targeted optimization of dynamic causal consistency, improving the physical plausibility of generated actions~\cite{vjepa2_score}.

\runhead{\textit{Aesthetic Reward ($R_{\text{aesthetic}}$)}}
The aesthetic reward is provided by a VLM. The VLM assesses each video for clarity, artifacts, and realism, assigning a discrete integer score (e.g., 1-5).

\runhead{\textit{GRPO Optimization}}
GRPO is an efficient, value-free policy optimization algorithm. At each optimization step, we sample a group of $G$ videos $\{x_0^i\}_{i=1}^G$ from the current policy $\pi_\theta$. The advantage $\hat{A}^i$ for each sample is computed by normalizing its reward relative to the group statistics: $\hat{A}^i = \bigl(R(x_0^i) - \text{mean}(\{R(x_0^j)\}_{j=1}^G)\bigr) / \text{std}(\{R(x_0^j)\}_{j=1}^G)$, where $\text{mean}(\cdot)$ and $\text{std}(\cdot)$ denote the group mean and standard deviation, respectively. This group-relative normalization provides a stable advantage estimation without requiring a separate critic network. The policy is then updated by maximizing the standard GRPO objective function:

\vspace{-1em}
\begin{small}
\begin{equation}
\begin{aligned}
&\mathcal{J}_{\text{GRPO}}(\theta) = \mathbb{E} \Bigg[ \frac{1}{G} \sum_{i=1}^{G} \bigg(
\min \big(r_i(\theta)\hat{A}^i, \\
&\text{clip}(r_i(\theta), 1-\epsilon, 1+\epsilon)\hat{A}^i \big) - \beta D_{\text{KL}}(\pi_\theta \| \pi_{\text{ref}}) \bigg) \Bigg].
\end{aligned}
\end{equation}
\end{small}
\vspace{-0.5em}

where $r_i(\theta) = \frac{\pi_\theta(x_0^i)}{\pi_{ref}(x_0^i)}$ is the importance sampling ratio, $\epsilon$ is a clipping hyperparameter that ensures conservative policy updates, and the KL-divergence term regularizes the policy towards the SFT policy $\pi_{\text{ref}}$ to mitigate reward hacking and maintain generation quality. This optimization process progressively aligns the MVG towards higher physical fidelity and aesthetic quality while maintaining its adherence to kinematic conditioning.

\vspace{-0.5em}
\section{Experiments}
\vspace{-0.5em}

\subsection{Experiment Settings}
\vspace{-0.25em}
\runhead{Architecture and Training}
The SRH employs Gemini-2.5 Pro~\cite{gemini25} as the core Vision-Language Model and Affordance-R1~\cite{affordance-r1} as the visual localizer. The MVG is initialized from the pretrained CogVideoX-5B~\cite{cogvideox} architecture. Experiments are conducted on the Bridge V2~\cite{bridge} dataset following the preprocessing protocol of~\cite{robomaster}, with a resolution of $480 \times 640$ pixels and 37 frames per subtask. Training proceeds in two stages: supervised fine-tuning (SFT) for 30,000 steps using AdamW with a learning rate of $2 \times 10^{-5}$, followed by GRPO post-training for 1,500 iterations at $5 \times 10^{-5}$. At inference time, generating a 111-frame long-horizon video spanning three subtasks requires approximately 50 GB of VRAM. Owing to its autoregressive subtask-based design, \MindV{} supports arbitrarily long task sequences with only linear growth in computational cost. All experiments are run on four NVIDIA H200 GPUs, with further implementation details provided in the Appendix.

\begin{figure*}[t]
  \centering
  \setlength{\abovecaptionskip}{-0.02em}   
   \includegraphics[width=1.0\linewidth]{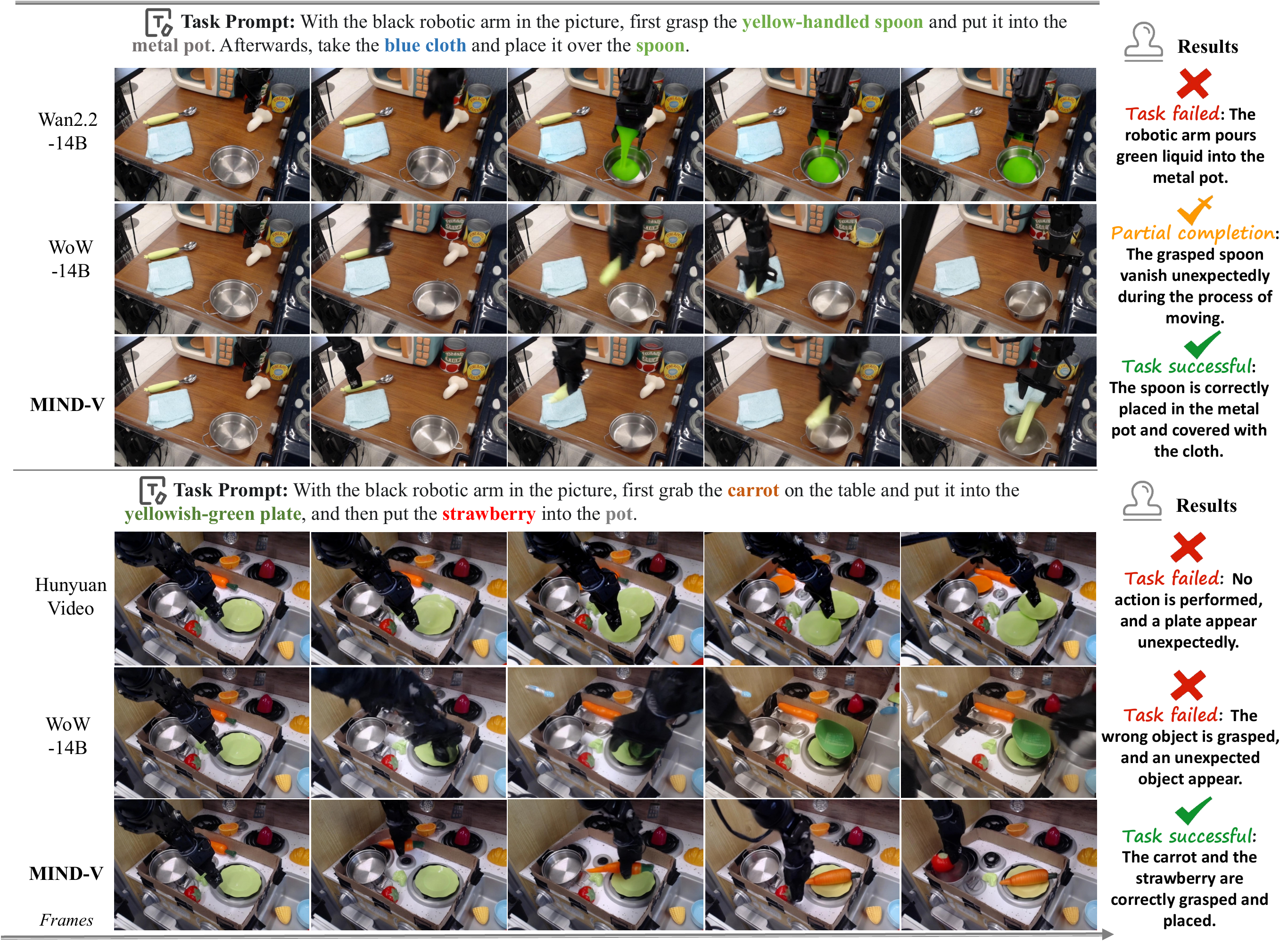}

  \caption{\textbf{Qualitative comparison of long-horizon robotic manipulation video generation.} }
\label{fig:qualitative_comparison}

   \label{fig:f4}
   \vspace{-1.5em}
\end{figure*}

\runhead{Evaluation Protocol and Metrics}
Evaluation is performed on a test set of 108 samples, comprising scenes from the Bridge V2 test set~\cite{bridge} and unseen web-sourced scenes. Given that different task horizons demand different criteria, we adopt a bifurcated evaluation protocol. For short-horizon tasks, we assess \textit{visual quality} using V-Bench~\cite{vbench2}. For long-horizon tasks, we additionally conduct a user study and report two metrics: (1) \textit{physical plausibility}, quantified by our Physical Foresight Coherence (PFC) score (Section~\ref{sec:GRPO}); and (2) \textit{task success rate}, defined as the average success rate across all subtasks of the full long-horizon task.

\begin{table*}[t]
\centering
\caption{\textbf{Visual quality evaluation on short-horizon and long-horizon tasks.} }
\label{tab:t1}
\vspace{-1.0em}
\scriptsize
\renewcommand{\arraystretch}{0.6}
\setlength{\tabcolsep}{3pt}

\resizebox{0.9\textwidth}{!}{%
\begin{tabular}{l|cccccc}
\toprule
\textbf{Method} & \makecell{Aesthetic\\Quality}$\uparrow$ & \makecell{Imaging\\Quality}$\uparrow$ & \makecell{Temporal\\Flicker}$\uparrow$ & \makecell{Motion\\Smoothness}$\uparrow$ & \makecell{Subject\\Consistency}$\uparrow$ & \makecell{Bg.\\Consistency}$\uparrow$ \\
\midrule
\multicolumn{7}{c}{\textbf{Short-horizon Tasks}} \\
\midrule
MotionCtrl~\cite{motionctrl}           & 0.491 & 0.665 & 0.977 & 0.972 & 0.915 & 0.942 \\
IRASim~\cite{IRASim}                    & 0.504 & 0.676 & 0.979 & 0.986 & 0.929 & 0.957 \\
Cosmos~\cite{cosmos}                    & 0.519 & 0.680 & 0.980 & \thirdbest{0.988} & 0.932 & 0.958 \\
DragAnything~\cite{draganything}        & 0.500 & 0.679 & {0.980} & 0.983 & {0.935} & 0.957 \\
Tora~\cite{tora}                        & 0.509 & 0.670 & 0.981 & 0.984 & 0.922 & \thirdbest{0.961} \\
RoboMaster~\cite{robomaster}            & 0.502 & \best{0.688} & \thirdbest{0.982} & 0.981 & \thirdbest{0.937} & 0.950 \\
Robodreamer~\cite{robodreamer}          & {0.511} & {0.680} & 0.977 & 0.976 & 0.930 & 0.945 \\
WoW-1-DiT-7B~\cite{wow}                 & \thirdbest{0.522} & 0.682 & \thirdbest{0.982} & {0.985} & 0.933 & {0.960} \\
\textbf{\MindV{} (Ours)}                  & \best{0.526} & \thirdbest{0.684} & \best{0.986} & \best{0.991} & \best{0.940} & \best{0.963} \\
\midrule
\multicolumn{7}{c}{\textbf{Long-horizon Tasks}} \\
\midrule
Robodreamer~\cite{robodreamer}          & 0.464 & 0.628 & 0.910 & 0.918 & 0.839 & 0.885 \\
WoW-1-DiT-7B~\cite{wow}                 & 0.476 & 0.635 & 0.922 & 0.929 & 0.851 & 0.894 \\
WoW-1-Wan-14B~\cite{wow}                & {0.498} & {0.652} & {0.935} & 0.950 & {0.874} & {0.906} \\
Dreamdojo~\cite{dreamdojo}                 & \thirdbest{0.504} & \best{0.660} & \thirdbest{0.944} & {0.948} & \thirdbest{0.883} & \thirdbest{0.909} \\
HunyuanVideo~\cite{hunyuanvideo}        & 0.487 & 0.643 & 0.928 & \thirdbest{0.952} & 0.862 & 0.900 \\
CogVideoX-5B~\cite{cogvideox}        & 0.484 & 0.640 & 0.926 & 0.947 & 0.860 & 0.895 \\
\textbf{\MindV{} (Ours)}                  & \best{0.512} & \thirdbest{0.658} & \best{0.955} & \best{0.953} & \best{0.896} & \best{0.924} \\
\bottomrule
\end{tabular}
}
\vspace{-2.0em}
\end{table*}

\runhead{Baselines and Comparative Setup}
For short-horizon tasks, we benchmark \MindV{} against two categories of baselines: autonomous world models (DreamDojo~\cite{dreamdojo}, RoboDreamer~\cite{robodreamer}, WoW~\cite{wow}, CogVideoX-5B~\cite{cogvideox}, HunyuanVideo~\cite{hunyuanvideo}) and trajectory-guided methods (IRASim~\cite{IRASim}, Cosmos~\cite{cosmos}, MotionCtrl~\cite{motionctrl}, DragAnything~\cite{draganything}, Tora~\cite{tora}). Video foundation models (CogVideoX-5B and HunyuanVideo) are fine-tuned on our dataset for fair comparison. Notably, trajectory-guided baselines receive privileged inputs at inference time, such as manual trajectories, masks, or anchor points, which are unavailable to our trajectory-free approach, underscoring \MindV{}'s ability to infer complex dynamics from high-level intent alone. For long-horizon tasks, which require complex planning and reasoning without explicit guidance, comparisons are restricted to the autonomous world model baselines. Each long-horizon task comprises 2 to 4 subtasks, designed to stress-test long-horizon planning and generation.

\vspace{-0.5em}
\subsection{Qualitative and Quantitative Comparison}
\vspace{-0.25em}
As shown in Fig.~\ref{fig:qualitative_comparison} and Tables~\ref{tab:t1}--\ref{tab:t2}, \MindV{} consistently outperforms all baselines across both short- and long-horizon tasks. On long-horizon benchmarks, \MindV{} surpasses the second-best method by 9.0\%, 76.7\%, and 172.2\% on PFC Score, Task Success Rate, and User Preference, respectively. Baseline models exhibit critical failure modes (Fig.~\ref{fig:qualitative_comparison}(a)), including logical hallucinations, physical implausibility (e.g., spontaneous object disappearance), and inaccurate semantic grounding (e.g., incorrect object manipulation), resulting in poor long-horizon coherence (Table~\ref{tab:t2}).

\MindV{} overcomes these limitations through its cognitive hierarchical design. The SRH and BSB collaborate to decompose user instructions into an explicit, executable plan, mitigating the semantic drift prevalent in end-to-end approaches. The MVG, aligned via the PFC reward, synthesizes videos that adhere to both physical laws and command requirements. Staged Visual Future Rollouts further suppress error accumulation, enabling coherent and physically plausible long-horizon manipulation sequences.

\begin{table}[t]
  \centering
  \small
  \renewcommand{\arraystretch}{0.95}
  \setlength{\tabcolsep}{3pt}
  \caption{Comprehensive evaluation of long-horizon tasks. Higher is better; best results are \textbf{bolded}.}
    \vspace{-1em}
  \label{tab:t2}
  \resizebox{\columnwidth}{!}{%
  \begin{tabular}{lccc}
  \toprule
  \multirow{2}{*}{\textbf{Method}} 
  & \parbox{1.7cm}{\centering PFC \\ Score~$\uparrow$} 
  & \parbox{1.9cm}{\centering Task \\ Success Rate } 
  & \parbox{1.7cm}{\centering User \\ Pref. (\%) $\uparrow$} \\
  \midrule
  Dreamdojo~\cite{dreamdojo}      & \thirdbest{0.424} & 0.333 & \thirdbest{18} \\
  Robodreamer~\cite{robodreamer}      & 0.418 & 0.275 & 7 \\
  WoW-1-DiT-7B~\cite{wow}             & 0.423 & 0.322 & 11 \\
  WoW-1-Wan-14B~\cite{wow}            & 0.420 & \thirdbest{0.347} & {14} \\
  CogVideoX-5B~\cite{cogvideox}             & 0.406 & 0.081 & 0 \\
  HunyuanVideo~\cite{hunyuanvideo}    & 0.411 & 0.098  & 1 \\
  \midrule
  \textbf{\MindV{} (Ours)}              & \best{0.462} & \best{0.613} & \best{49} \\
  \bottomrule
  \end{tabular}%
  }
  \vspace{-1em}
\end{table}

\begin{figure}[t]
  \centering
   \setlength{\abovecaptionskip}{-0.02em}   
  \includegraphics[width=\columnwidth]{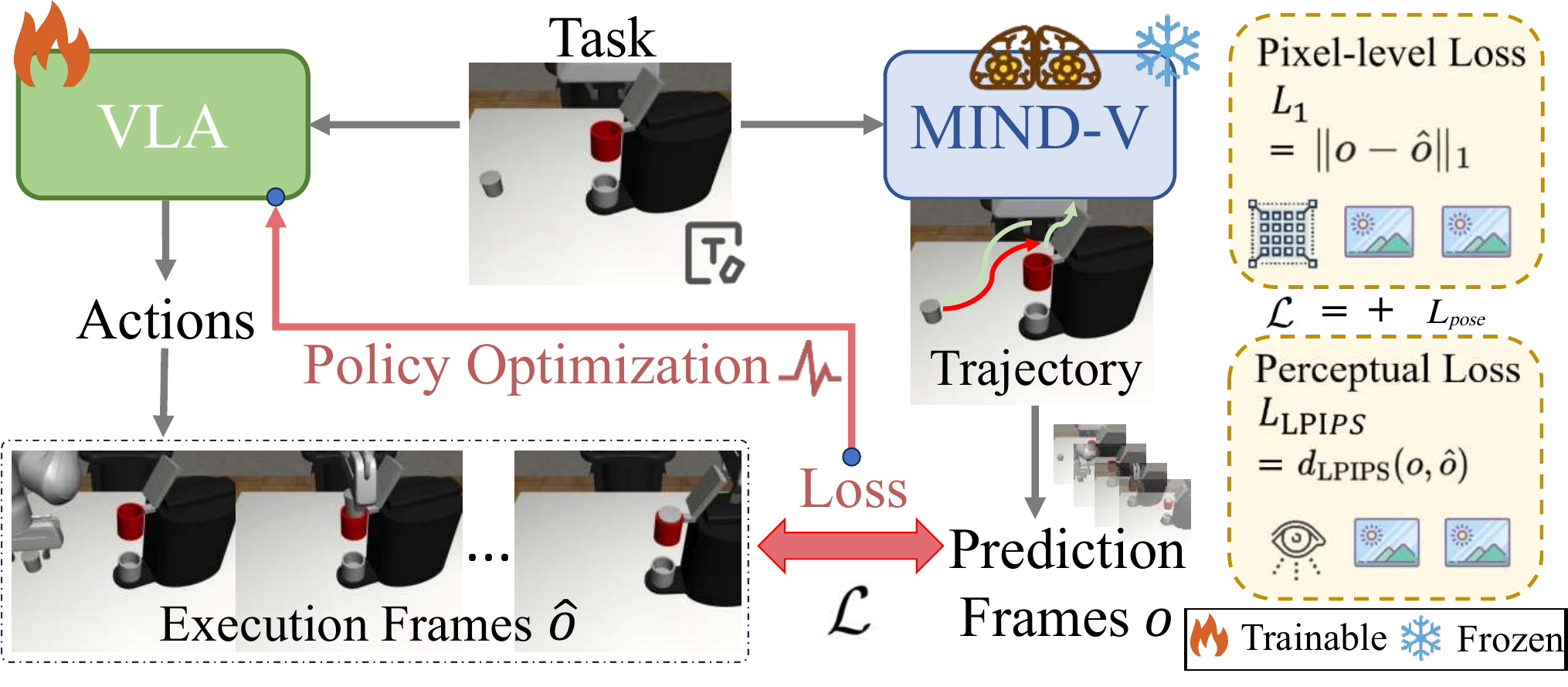}
  \caption{\textbf{Policy learning with \MindV{}.}}
  \label{fig:f6}
  \vspace{-1em}
\end{figure}

\vspace{-0.5em}
\subsection{World Model for Policy Learning}
\vspace{-0.25em}
To validate \MindV{} as a training ground for embodied agents, we conduct a generation-to-policy experiment on the MimicGen simulation benchmark~\cite{mimicgen} (Fig.~\ref{fig:f6}), evaluating three fine-grained manipulation tasks (Coffee, StackThree, Square) with 128 demonstration trajectories per task.

\runhead{Setup}
We first fine-tune OpenVLA-OFT~\cite{openvla} via imitation learning (IL) on 300 expert trajectories per task to obtain a base policy. The IL baseline continues training this model under end-effector pose supervision ($\mathcal{L}_{\text{pose}}$) until convergence. Our approach augments the pose loss with a visual goal derived from \MindV{}: given a task instruction, \MindV{} generates a successful visual rollout $o$, against which the VLA's predicted rollout $\hat{o}$ is supervised:
\vspace{-0.75em}
\begin{equation}
\mathcal{L} = \lambda_{\text{pix}} \|\hat{o} - o\|_1 + \lambda_{\text{perc}} d_{\text{LPIPS}}(\hat{o}, o) + \lambda_{\text{IL}}\mathcal{L}_{\text{pose}},
\end{equation}
where $\|\hat{o} - o\|_1$ and $d_{\text{LPIPS}}(\hat{o}, o)$ denote the $L_1$ pixel reconstruction loss and LPIPS perceptual loss~\cite{LPIPS}, respectively.

\runhead{Discussion}
As shown in Table~\ref{tab:policy_learning}, \MindV{} guidance raises the mean success rate from 27.7\% (Base) to \textbf{43.5\%}, substantially exceeding the 33.4\% achieved by IL continuous training. This result highlights a key advantage of pixel-space world models: operating in the same visual space as the VLA's pre-training data, \MindV{}'s imagined rollouts directly engage the model's visual priors, effectively bridging generative world simulation and actionable policy learning~\cite{WMPO}.

\begin{table}[t]
  \centering
  \footnotesize
  \renewcommand{\arraystretch}{1.1} 
  \setlength{\tabcolsep}{3pt} 

  \caption{Task success rates (\%) under different training paradigms in the MimicGen simulation benchmark.}
    \vspace{-1em}
  \label{tab:policy_learning}
  \resizebox{0.75\columnwidth}{!}{%
  \begin{tabular}{lcccc} 
  \toprule
  \textbf{Methods} & \textbf{Coffee} & \textbf{StackThree} & \textbf{Square} & \textbf{Mean} \\
  \midrule
  Base policy & 32.6 & 30.2 & 20.2 & 27.7 \\
  \midrule 
  IL          & 37.4 & 36.7 & 26.1 & 33.4 \\
  \textbf{\MindV{} + IL} & \textbf{51.7} & \textbf{48.3} & \textbf{30.4} & \textbf{43.5} \\
  \bottomrule
  \end{tabular}%
  }
  \vspace{-1.5em}
\end{table}

\vspace{-0.5em}
\subsection{Ablation Study}
\label{ablation}
\vspace{-0.5em}
\runhead{Component Contribution Analysis}
To evaluate the contribution of each core component, we compare the full \MindV{} model against four ablated variants on long-horizon simulation: (a) \textbf{w/o GRPO}: trained solely with SFT, without GRPO post-training stage; (b) \textbf{Re Affordance w/ YOLO+SAM2}: replacing the affordance localizer~\cite{affordance-r1} in BSB with a YOLO-World~\cite{YOLOWorld} + SAM2~\cite{sam2} pipeline; (c) \textbf{w/o Staged Rollouts}: disabling the test-time staged optimization mechanism; (d) \textbf{Re Gemini w/ Qwen3-VL}: substituting Gemini 2.5 Pro in SRH with Qwen3-VL~\cite{qwen3} to test VLM dependence.

\begin{table}[t]
\centering
\footnotesize
\renewcommand{\arraystretch}{0.7}
\setlength{\tabcolsep}{5pt}
\caption{Ablation study on long-horizon simulation tasks. Visual quality (Aesthetic and Imaging Quality) and functional correctness (PFC Score and subtask Average Success Rate) are reported. Best results are \textbf{bolded}.}
\vspace{-1.0em}
\label{tab:ablation_study}
\resizebox{0.5\textwidth}{!}{%
\begin{tabular}{lcccc}
\toprule
\multirow{2}{*}{\textbf{Model Variant}} & \multicolumn{2}{c}{\textbf{Visual Quality}} & \multicolumn{2}{c}{\textbf{Functional Correctness}} \\
\cmidrule(lr){2-3} \cmidrule(lr){4-5}
& Aesthetic~$\uparrow$ & Imaging~$\uparrow$ & PFC Score~$\uparrow$ & subtask Avg.~$\uparrow$ \\
\midrule
(a) w/o GRPO                  & 0.491 & 0.675 & 0.429 & 0.582 \\
(b) Re Affordance w/ YOLO & 0.498 & 0.680 & 0.445 & 0.455 \\
(c) w/o Staged Rollouts       & 0.482 & 0.671 & 0.438 & 0.327 \\
(d) Re Gemini w/ Qwen3-VL     & 0.500 & 0.679 & 0.452 & 0.567 \\
\midrule
\textbf{\MindV{} (Full)}        & \textbf{0.504} & \textbf{0.684} & \textbf{0.462} & \textbf{0.613} \\
\bottomrule
\end{tabular}%
}
\vspace{-0.75em}
\end{table}

As shown in Table~\ref{tab:ablation_study}, the full model outperforms all variants, confirming the necessity of each component. Removing GRPO yields a clear decline in PFC Score, underscoring the effectiveness of RL-based alignment for physical plausibility. Replacing the affordance module causes a substantial drop in subtask success rate, highlighting its critical role in functional grounding within the BSB. Disabling staged rollouts leads to pronounced degradation, demonstrating that the propose-verify-refine mechanism is essential for mitigating error accumulation over extended horizons. Substituting the VLM results in only a marginal performance drop, indicating that \MindV{} benefits from but does not depend on a specific VLM.

\begin{table}[t]
  \centering
  \footnotesize
  \renewcommand{\arraystretch}{0.95}
  \setlength{\tabcolsep}{3pt}
  \caption{Sensitivity analysis of reward weights.}
    \vspace{-1em}
  \label{tab:sensitivity}
  \resizebox{0.95\columnwidth}{!}{%
  \begin{tabular}{cc|cccc}
  \toprule
  $w_a$ & $w_p$ & Aesthetic $\uparrow$ & Imaging $\uparrow$ & PFC Score $\uparrow$ & subtask Avg. $\uparrow$ \\
  \midrule
  0.5   & 0.2   & 0.498 & 0.651 & \textbf{0.449} & 0.612 \\
  \textbf{1.0} & \textbf{0.2} & \underline{0.504} & \textbf{0.658} & \underline{0.445} & \textbf{0.613} \\
  2.0   & 0.2   & \textbf{0.508} & 0.657 & 0.434 & 0.562 \\
  \bottomrule
  \end{tabular}%
  }
  \vspace{-1.5em}
\end{table}

\runhead{Reward Sensitivity Analysis}
These parameters $w_a$ and $w_p$ in Eq.~\ref{eq:reward} serve to normalize reward scales to the same order of magnitude. The sensitivity analysis (Table~\ref{tab:sensitivity}) on long-horizon tasks shows robust performance across a reasonable range and the rationality of our parameter selection.



\vspace{-0.5em}
\section{Conclusion}
\vspace{-0.5em}
We presented \MindV{}, a cognitive hierarchical video world model for simulating the physical dynamics of long-horizon robotic manipulation. Inspired by human cognitive pathways, the model integrates a Semantic Reasoning Hub, a Behavioral Semantic Bridge, and a Motor Video Generator. This decoupled architecture separates high-level semantic reasoning from low-level pixel synthesis, ensuring long-horizon coherence, semantic grounding, and physical plausibility. To strictly enforce physical laws and prevent error accumulation, we introduced a GRPO post-training phase guided by a Physical Foresight Coherence reward alongside a test-time optimization strategy utilizing staged visual rollouts. Extensive evaluations demonstrate that \MindV{} achieves state-of-the-art video synthesis and establishes a highly effective simulation environment for training embodied AI policies.

\section{Limitations}
\label{sec:limitations}
Despite generating physically compliant and semantically consistent manipulation videos, \MindV{} exhibits two primary limitations. First, the Staged Visual Future Rollouts mechanism incurs additional inference overhead, as the propose-verify-refine cycle generates multiple candidate videos per subtask transition. Nonetheless, this fully automated process offsets the cost by eliminating the extensive random sampling and manual cherry-picking required by baseline methods. Second, the framework is sensitive to upstream SRH inaccuracies, where affordance localization failures~\cite{affordance-r1} can propagate downstream. To mitigate this, we introduce a VLM-based fallback mechanism that infers target coordinates and destinations via semantic and spatial reasoning. These coordinates are then refined and applied as point prompts for a segmentation model~\cite{sam2} to extract precise masks.

\bibliography{main}

\clearpage
\appendix
\section{Analysis of Computational Cost and Hyperparameters}
\label{sec:supp_cost_analysis}

This section provides a detailed analysis of the computational cost and key hyperparameters of the \MindV{} framework. We first examine the world model's scalability with respect to task length and then present an ablation study on the number of staged visual future rollouts (\textit{K}) utilized in our test-time optimization strategy.

\subsection{Scalability with Task Length}

\begin{figure}[t]
    \centering
    \includegraphics[width=\linewidth]{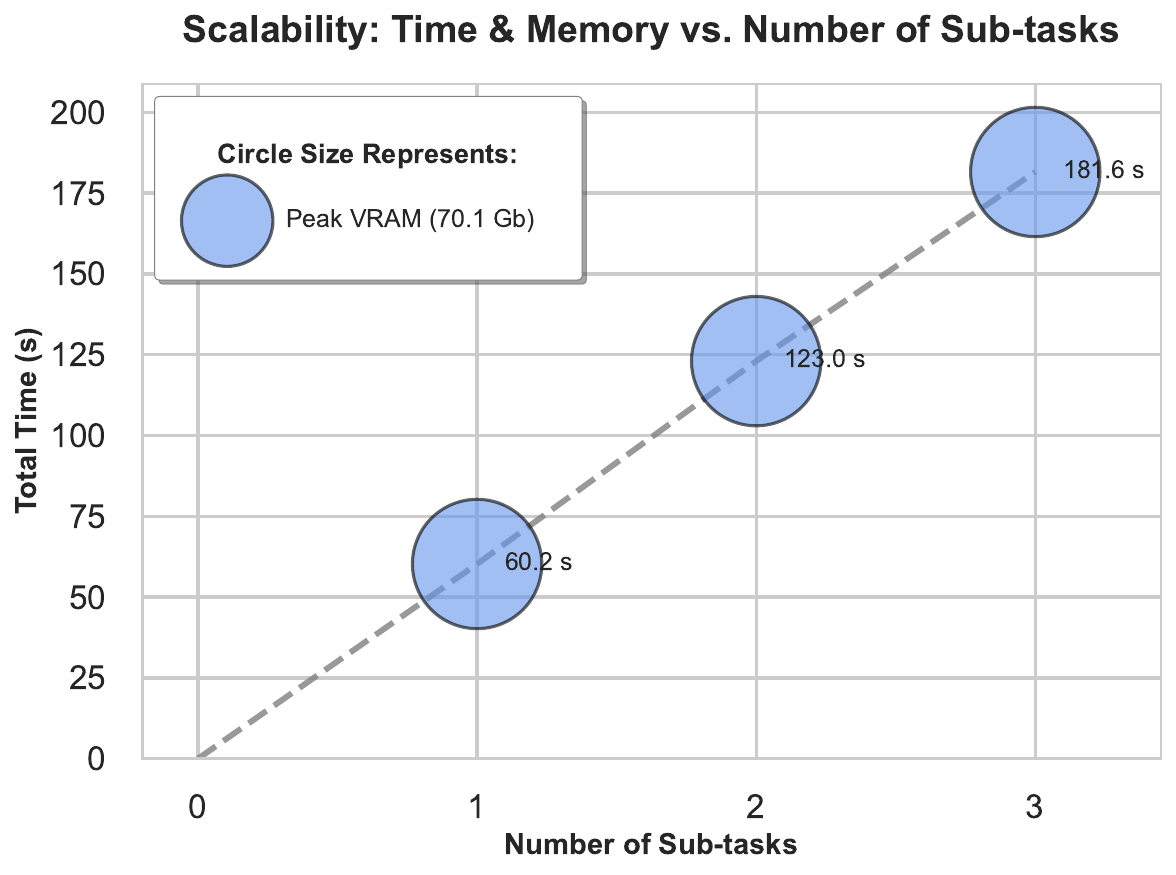}
    \caption{\textbf{Scalability of \MindV{}.} Total generation time (Y-axis) scales linearly with the number of sub-tasks (X-axis). Circle size represents peak VRAM, which remains constant, demonstrating the memory efficiency of our approach.}
    \label{fig:supp_scalability}
    \vspace{-1.5em}
\end{figure}

To validate \MindV{}'s efficiency as a world simulator in long-horizon tasks, we measure the simulation time and peak VRAM usage across tasks ranging from one to three sub-tasks. The key metrics are summarized in Table~\ref{tab:cost_analysis} and visualized in Figure~\ref{fig:supp_scalability}.

Our findings highlight two critical properties of the proposed architecture. First, the total simulation time scales linearly with the number of sub-tasks. The average time per sub-task remains constant at approximately 60  seconds \footnote{This is a reference value using the Gemini-2.5 API~\cite{gemini25}. The inference speed of the SRH is influenced by multiple factors, including VLM API response time and network latency. Reported times represent pure inference duration, excluding network transmission latency.}, demonstrating that our framework's computational time scales predictably with task length. Second, and crucially, the peak VRAM usage remains constant regardless of  the task length. As shown by the consistently sized circles in Figure~\ref{fig:supp_scalability}, the peak VRAM remains constant at approximately 70 GB. This constant memory footprint is a direct benefit of our hierarchical and autoregressive design, where memory is allocated for a single sub-task and subsequently reused, enabling highly scalable visual world simulation for long task sequences.

\begin{table*}[h]
\centering
\caption{\textbf{Computational cost as a function of task length.} We report total time, average time per sub-task, peak VRAM usage, and the percentage distribution of time and VRAM across the SRH planning and MVG simulation stages.}
\label{tab:cost_analysis}
\resizebox{\textwidth}{!}{%
\begin{tabular}{c|ccc|cc|cc}
\toprule
\textbf{No. of} & \textbf{Total} & \textbf{Avg. Time per} & \textbf{Peak VRAM} & \multicolumn{2}{c|}{\textbf{Time Dist. (\%)}} & \multicolumn{2}{c}{\textbf{VRAM Dist. (\%)}} \\
\cmidrule(lr){5-6} \cmidrule(lr){7-8}
\textbf{Sub-tasks} &  \textbf{Time (s)} & \textbf{Sub-task (s)} & \textbf{(GB)} & \textbf{Plan (SRH)} & \textbf{Gen (MVG)} & \textbf{Plan (SRH)} & \textbf{Gen (MVG)} \\
\midrule
1 & 60.24 & 30.14 & 70.12 & 36.5\% & 63.5\% & 14.1\% & 85.9\% \\
2 & 123.02 & 30.60 & 70.12 & 34.3\% & 65.7\% & 14.4\% & 85.6\% \\
3 & 181.55 & 30.85 & 70.12 & 32.4\% & 67.6\% & 14.0\% & 86.0\% \\
\bottomrule
\end{tabular}%
}
\end{table*}

Furthermore, Table~\ref{tab:cost_analysis} details the internal resource distribution. The Motor Video Generator (MVG) constitutes the primary computational bottleneck, consuming the majority of both execution time (approximately 65-70\%) and VRAM (approximately 86\%). In contrast, the Semantic Reasoning Hub (SRH) planning stage introduces only a minor and stable computational overhead.

\subsection{Analysis of the Number of Rollout Samples (K)}

The Staged Visual Future Rollouts mechanism is governed by a key hyperparameter, \textit{K}, which defines the number of candidate future trajectories simulated at each sub-task transition. While a larger \textit{K} increases the probability of identifying a physically plausible future, it also incurs greater computational cost. To analyze this trade-off, we conduct an ablation study by varying \textit{K} from 1 to 5.

\begin{figure*}[t!]
    \centering
    \includegraphics[width=0.45\linewidth]{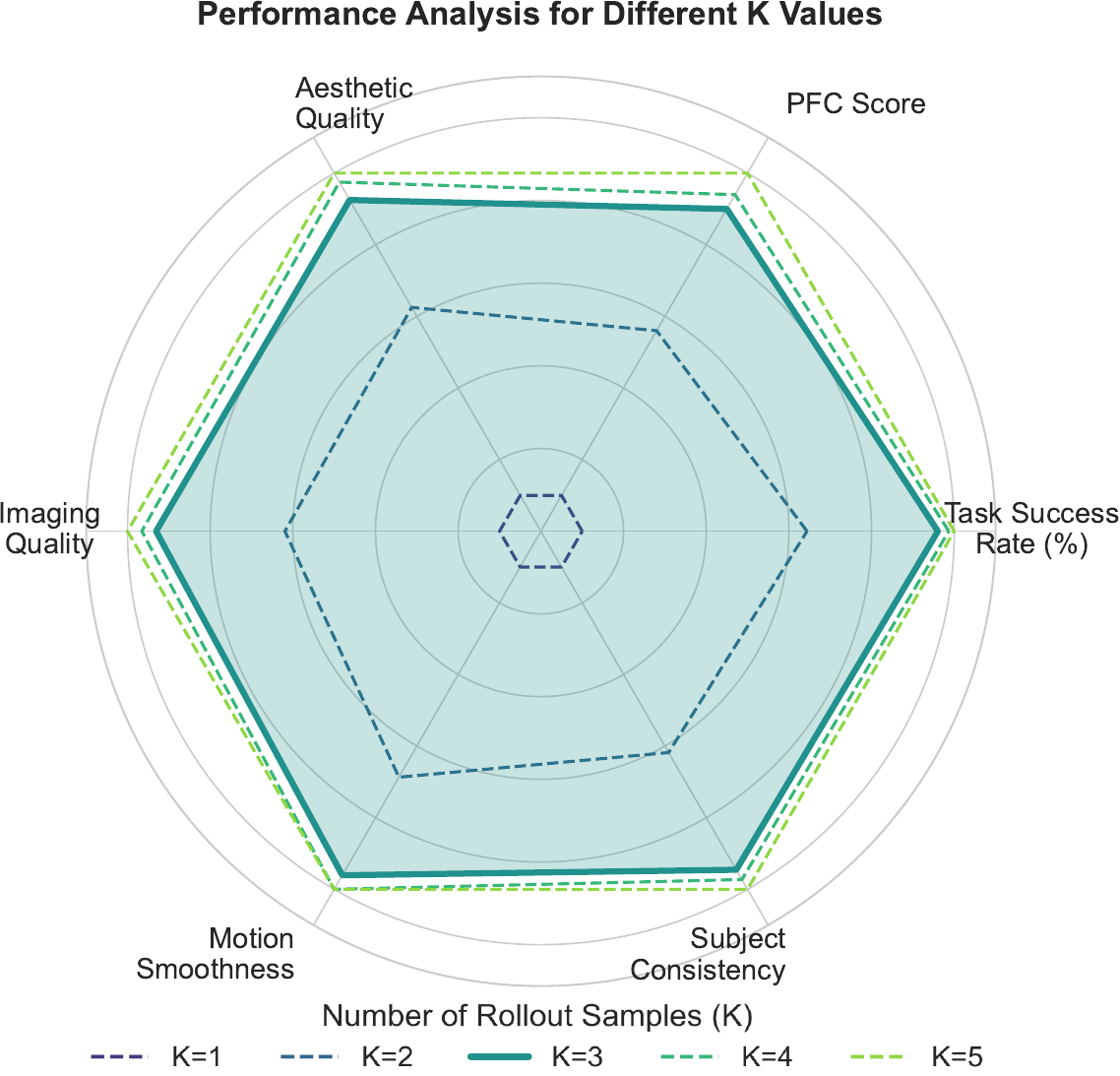} 
    \hfill 
    \includegraphics[width=0.53\linewidth]{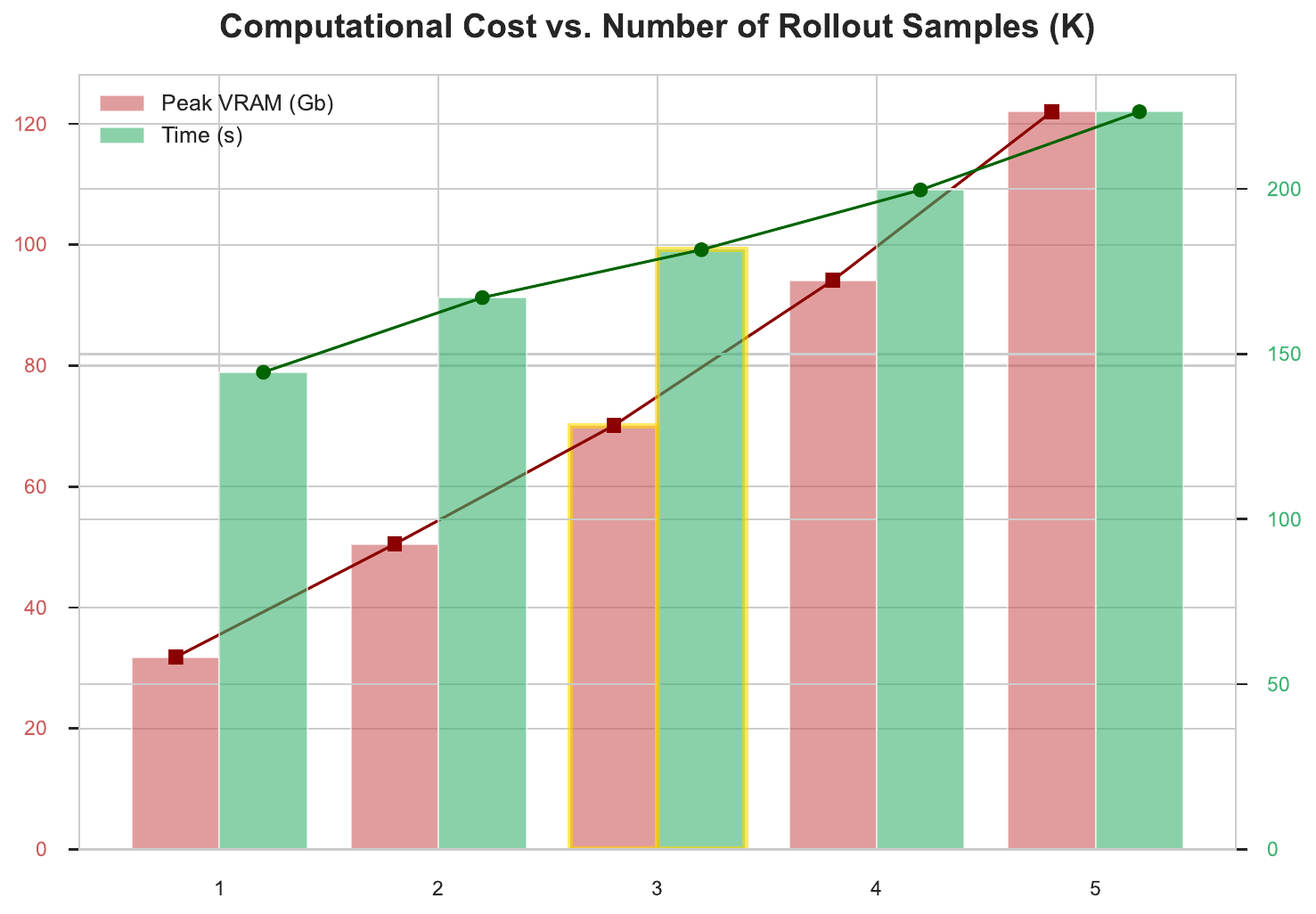} 
    \caption{\textbf{Analysis of the trade-off for the number of visual future rollouts (\textit{K})}. \textbf{(Left)} The performance radar chart shows that the overall performance area expands significantly up to \textit{K}=3 but exhibits diminishing returns thereafter. \textbf{(Right)} The cost chart shows that both time and Peak VRAM increase steadily with \textit{K}, with memory cost escalating significantly. \textit{K}=3 (highlighted) is chosen as the optimal balance.}
    \label{fig:supp_k_analysis}
\end{figure*}

The results, visualized in Figure~\ref{fig:supp_k_analysis}, illustrate the relationship between simulation fidelity and computational cost. As shown in the performance radar chart (Figure~\ref{fig:supp_k_analysis}, left), a significant performance uplift is observed as \textit{K} increases from 1 to 3, demonstrating the effectiveness of the rollout mechanism in filtering out physically incoherent futures. For instance, the Task Success Rate increases from 35.2\% at \textit{K}=1 to 61.3\% at \textit{K}=3. However, this trend exhibits sharply diminishing returns, with only marginal gains when increasing \textit{K} from 3 to 5. In contrast, the computational cost scales unfavorably with larger \textit{K} as shown in the bar chart (Figure~\ref{fig:supp_k_analysis}, right). For example, Peak VRAM consumption nearly doubles from 70.1 GB at \textit{K}=3 to 122.0 GB at \textit{K}=5. This analysis confirms that \textit{K}=3 strikes an optimal balance between simulation accuracy and computational efficiency. Therefore, we adopt \textit{K}=3 as the default setting for all experiments.

\begin{table*}[h!]
\centering
\caption{\textbf{Ablation study on the number of visual future rollouts (\textbf{\textit{K}}).} We evaluate the impact of varying \textit{K} on functional correctness, visual quality, and computational cost. The setting \textit{K}=3 (highlighted row) achieves the best balance between performance and efficiency.}
\label{tab:ablation_k}
\resizebox{\textwidth}{!}{%
\begin{tabular}{c|cc|cccccc}
\toprule
\multirow{2}{*}{\textit{K}} & \multicolumn{2}{c|}{\textbf{Cost}} & \multicolumn{6}{c}{\textbf{Performance}} \\
\cmidrule(lr){2-3} \cmidrule(lr){4-9}
& \makecell{Time (s) \\ $\downarrow$} & \makecell{Peak VRAM (GB) \\ $\downarrow$} & \makecell{Task Success \\ Rate (\%)~$\uparrow$} & \makecell{PFC Score \\ $\uparrow$} & \makecell{Aesthetic \\ Quality~$\uparrow$} & \makecell{Imaging \\ Quality~$\uparrow$} & \makecell{Motion \\ Smoothness~$\uparrow$} & \makecell{Subject \\ Consistency~$\uparrow$} \\
\midrule
1 & 144.5  & 31.8 & 35.2 & 0.405 & 0.471 & 0.660 & 0.931 & 0.865 \\
2 & 167.1  & 50.5 & 51.7 & 0.428 & 0.492 & 0.675 & 0.946 & 0.884 \\
\rowcolor{gray!20}
\textbf{3} & \textbf{181.6} & \textbf{70.1} & \textbf{61.3} & \textbf{0.445} & \textbf{0.504} & \textbf{0.684} & \textbf{0.953} & \textbf{0.896} \\
4 & 199.7  & 94.1 & 62.1 & 0.447 & 0.506 & 0.685 & 0.954 & 0.897 \\
5 & 223.4 & 122.0 & 62.5 & 0.450 & 0.507 & 0.686 & 0.954 & 0.898 \\
\bottomrule
\end{tabular}%
}
\end{table*}

\section{Dataset Construction}
\label{sec:supp_dataset}

\begin{figure*}[t!]
    \centering
    \includegraphics[width=\linewidth]{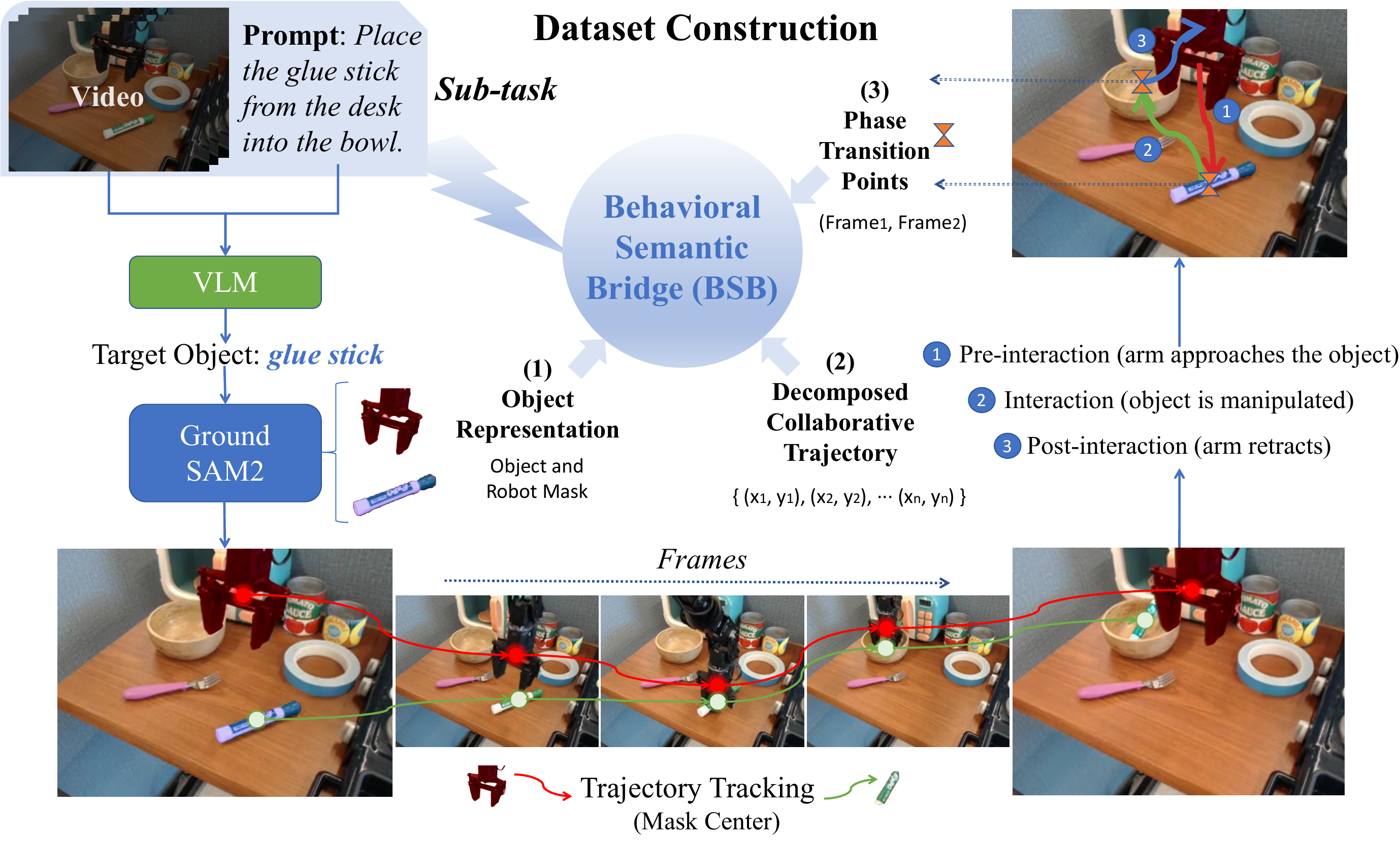} 
    \caption{\textbf{Overview of our automated BSB annotation pipeline.} A VLM first extracts the target object from the language prompt, which is then used by Grounded SAM2~\cite{groundsam} to generate the (1) Object Representation (masks). Concurrently, trajectory tracking is performed on the object and gripper masks. The trajectory is partitioned based on the object's motion to produce the (2) Decomposed Collaborative Trajectory and (3) Phase Transition Points. These components collectively form the structured BSB annotation used for SFT.}
    \label{fig:dataset}
\end{figure*}

The Supervised Fine-Tuning (SFT) stage of our training paradigm requires a large-scale dataset of robotic manipulation videos annotated with our structured Behavioral Semantic Bridge (BSB) representation. To this end, we developed an automated pipeline to generate ground-truth BSB annotations from the raw Bridge V2 dataset~\cite{bridge} following the data processing protocol established in~\cite{robomaster}, as illustrated in Figure~\ref{fig:dataset}. This pipeline comprises two primary stages: Object Representation Generation and Trajectory Decomposition.

\begin{figure*}[t!]
    \centering
    \includegraphics[width=\linewidth]{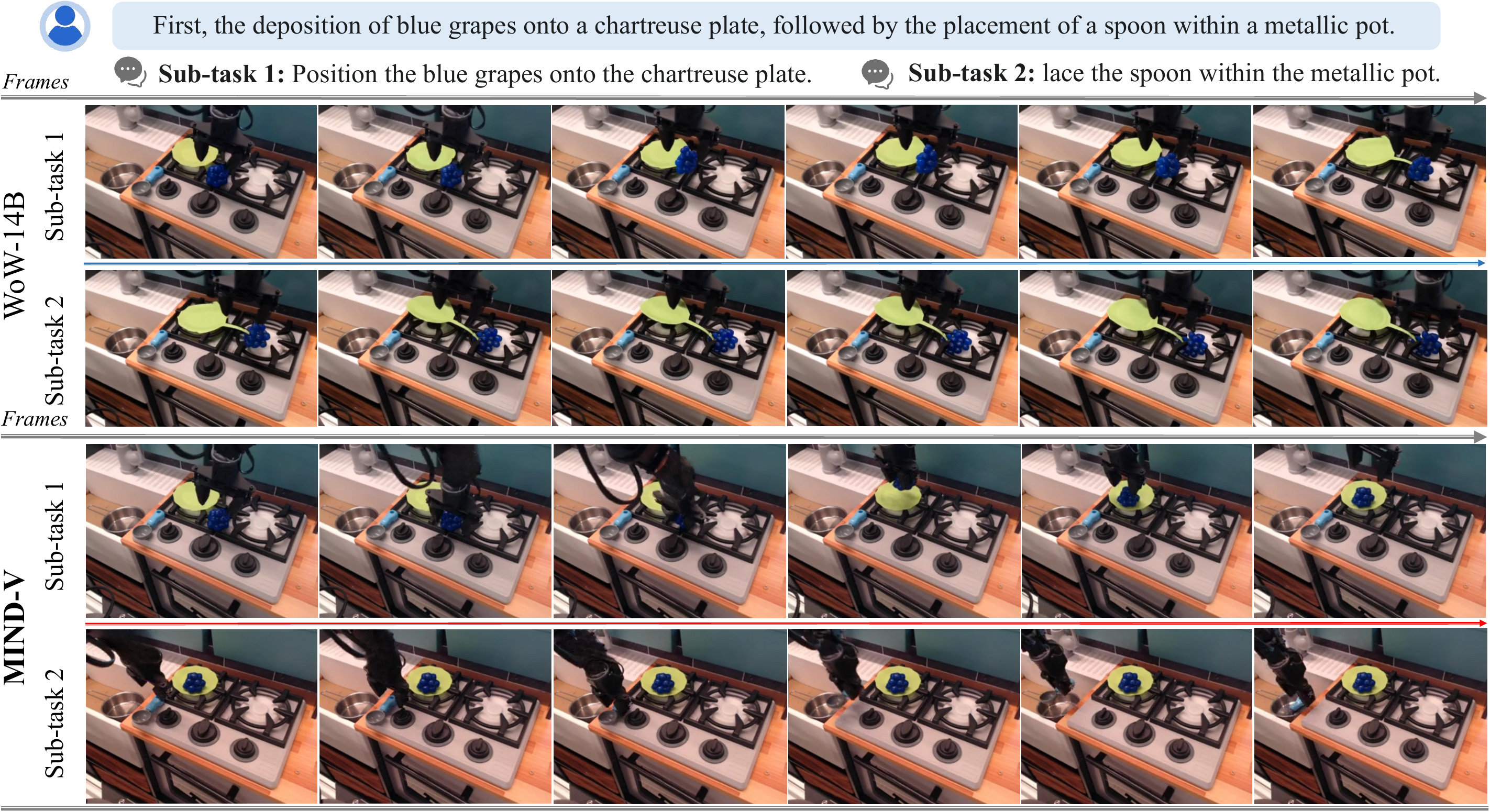} 
     \caption{\textbf{Qualitative comparison on a complex long-horizon task.} The model is instructed to first place blue grapes onto a chartreuse plate, and then place a spoon into a metallic pot. \textbf{(Top)} The baseline model, WoW-14B~\cite{wow}, exhibits a catastrophic failure in long-horizon reasoning. In Sub-task 1, the grapes levitate without being touched, a clear physical violation. In Sub-task 2, it demonstrates severe semantic grounding error by incorrectly interacting with the plate instead of the instructed spoon, resulting in a complete breakdown of logical coherence. \textbf{(Bottom)} In stark contrast, \MindV{} successfully executes the full sequence, correctly completing both sub-tasks as instructed. This result validates the efficacy of our hierarchical architecture; the SRH's explicit planning and the BSB's structured guidance prevent the semantic drift and error accumulation that plague the baseline, ensuring robust execution of multi-step instructions.}
    \label{fig:vis_long}
\end{figure*}

\subsection{Object Representation Generation}
This stage derives the \textbf{Object Representation} (segmentation masks) for both the manipulated object ($M_{\text{obj}}$) and the robot arm ($M_{\text{rob}}$) by visually grounding the natural language instructions. For each sub-task video and its corresponding instruction (e.g., “pick up the red block”), the extraction process proceeds as follows:

\begin{enumerate}
    \item \textbf{\textit{Object Identification:}} A pre-trained Vision-Language Model (VLM), such as Qwen-VL 2.5~\cite{Qwen-VL}, extracts the noun phrase corresponding to the object of manipulation (e.g., “red block”) from the instruction.
    \item \textbf{\textit{Language-Grounded Segmentation:}} The extracted noun phrase serves as a text prompt for Grounded SAM2~\cite{groundsam}, an open-vocabulary segmentation model, which generates the pixel-wise segmentation mask for the target object ($M_{\text{obj}}$) in the initial frame.
    \item \textbf{\textit{Robot Arm Mask:}} Because the manipulator's visual appearance remains consistent across the dataset, we apply a pre-defined template mask to obtain the robot arm segmentation ($M_{\text{rob}}$).
\end{enumerate}

We reserve a randomly sampled 10\% of the data for testing and use the remainder for training.

\subsection{Trajectory Decomposition and Phase Segmentation}

This stage tracks and partitions the trajectories of the robot and the object into three meaningful phases: pre-interaction, interaction, and post-interaction. The process is based on the motion state of the manipulated object.

We first employ a video object tracking model, in this case SAM2~\cite{sam2}, to track the segmentation masks of both the target object and the robot gripper throughout the video sequence. The center point of these masks forms the raw trajectory data. The \textbf{Decomposed Collaborative Trajectory} is then segmented based on the object's motion:
\begin{enumerate}
    \item \textbf{\textit{Pre-interaction Phase ($T_{\text{pre}}$):}} This phase is defined as the sequence of frames from the start of the sub-task until the target object begins to move.
    \item \textbf{\textit{Interaction Phase ($T_{\text{interact}}$):}} This phase covers the frames during which the target object is in motion.
    \item \textbf{\textit{Post-interaction Phase ($T_{\text{post}}$):}} This phase begins once the target object comes to rest again and continues until the end of the sub-task.
\end{enumerate}

The spatiotemporal trajectories of the robot arm and the manipulated object are determined by the paths of their respective mask centroids during these phases. The \textbf{Phase Transition Points} ($F_{\text{pre}}, F_{\text{interact}}, F_{\text{post}}$) are defined by the start and end points of the object's motion. To ensure fidelity of world dynamics, failure cases from the automated pipeline, such as incorrect grounding or trajectory errors, are flagged for manual correction by human annotators.

\begin{figure*}[t!]
    \centering
    \includegraphics[width=\linewidth]{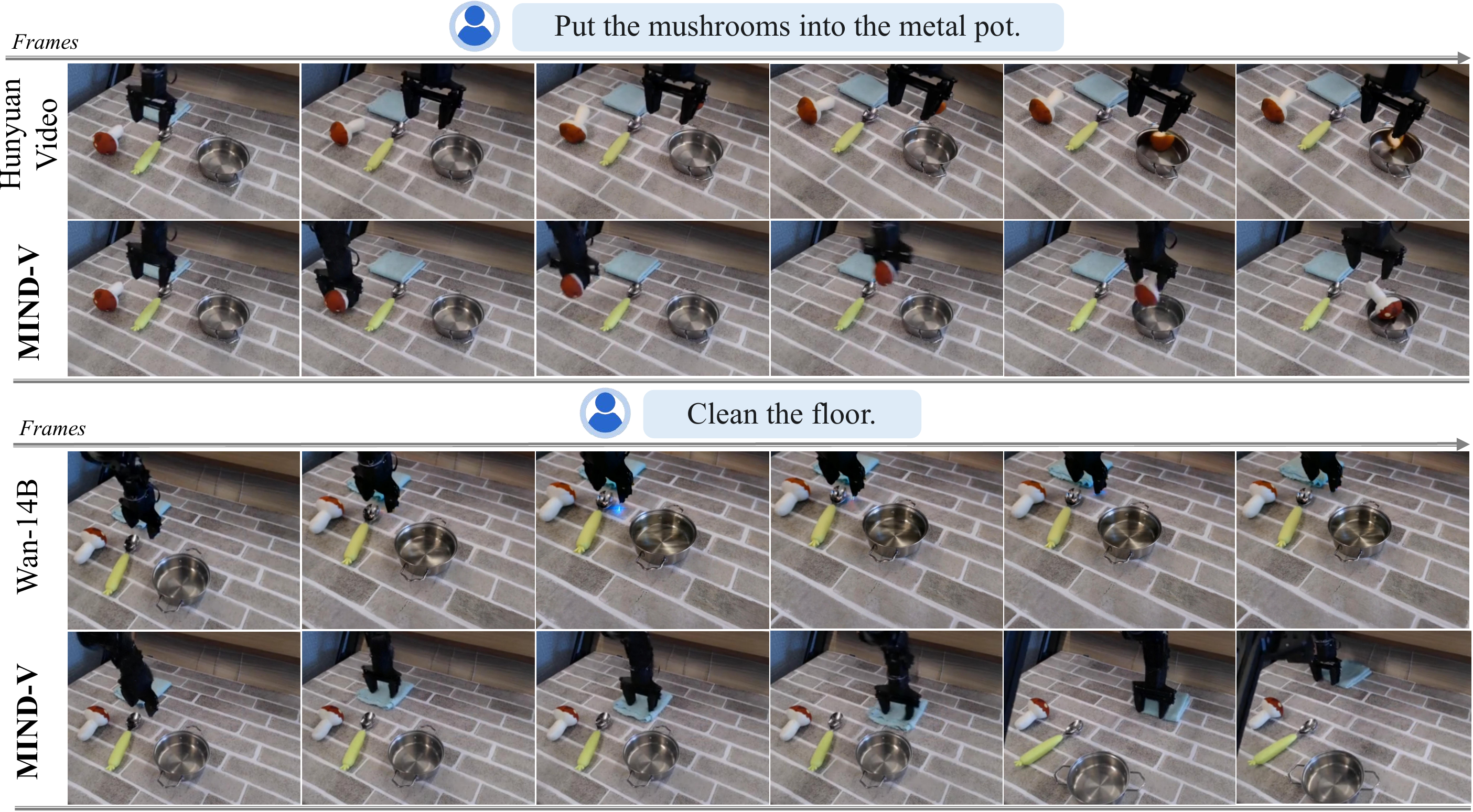} 
    \caption{\textbf{Qualitative comparison on short-horizon tasks.} This figure illustrates performance on two distinct single-step instructions. \textbf{(Top)} For “Put the mushrooms into the metal pot”, the baseline (HunyuanVideo~\cite{hunyuanvideo}) exhibits physical implausibility, with the mushroom clipping through the pot's rim. \MindV{}, in contrast, simulates a physically plausible interaction. \textbf{(Bottom)} For the more abstract instruction “Clean the floor”, the baseline (Wan-14B~\cite{wan22}) fails to take any action, demonstrating a lack of semantic grounding. \MindV{} correctly interprets the instruction, grasps the cloth, and performs a wiping motion, showcasing its superior planning and reasoning capabilities.}
    \label{fig:vis_short}
\end{figure*}

\section{Additional Visual Results}
\label{sec:supp_visuals}

\begin{figure*}[t!]
    \centering
    \includegraphics[width=\linewidth]{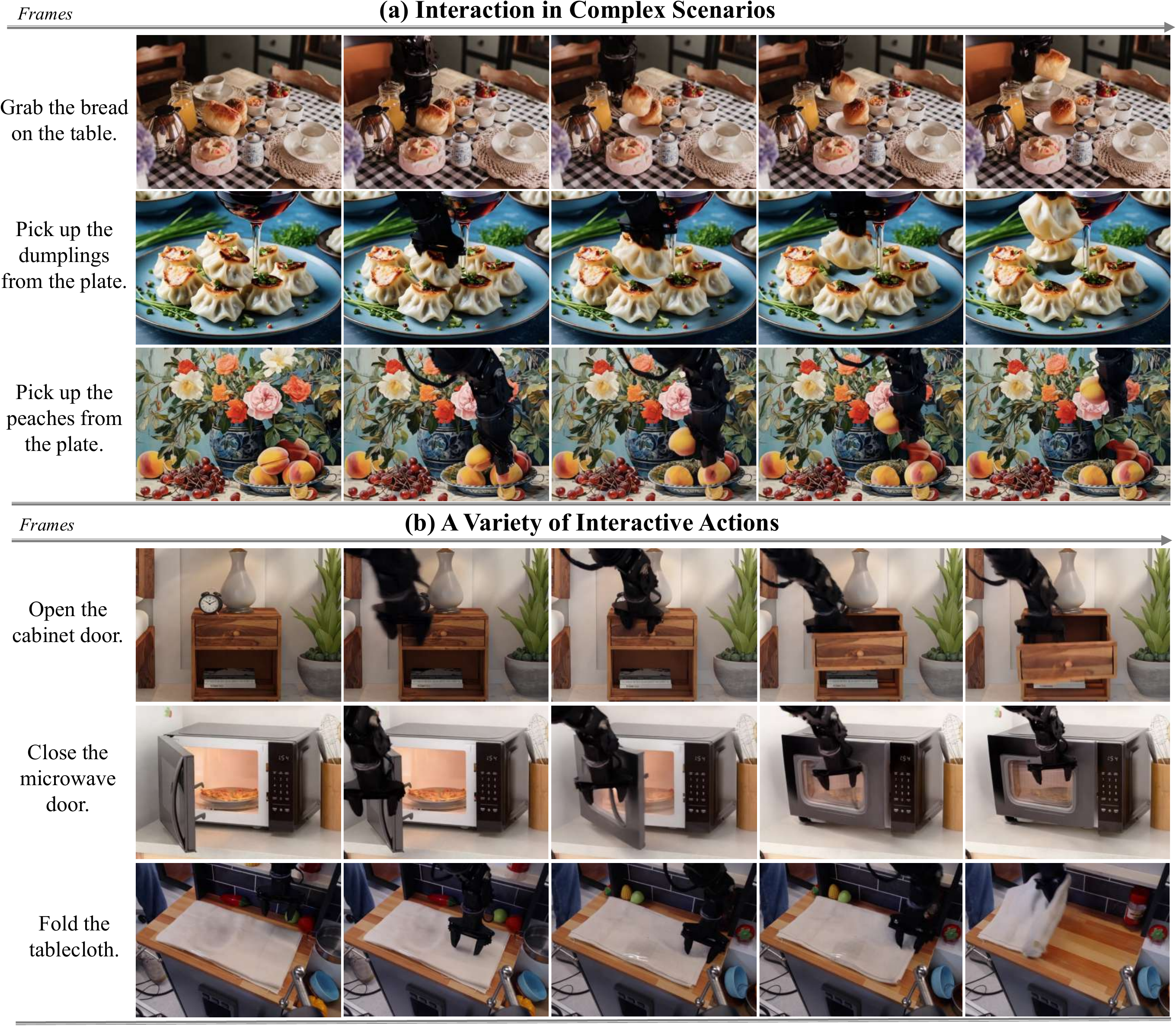} 
    \caption{\textbf{Generalization to complex scenarios and diverse manipulation skills.} Panel (a) demonstrates the robust generalization ability of \MindV{} in out-of-distribution (OOD) scenarios. The world model accurately isolates and manipulates targets in cluttered environments (e.g., grabbing bread from a full table or picking a dumpling), as well as in stylistically distinct scenes (e.g., picking peaches in an artistic setting). The domain-invariant BSB representation ensures precise control without disturbing the background fidelity. Panel (b) highlights the simulation of diverse interactive dynamics where the model leverages affordance-aware reasoning to execute physics-compliant interactions. This includes manipulating articulated objects like opening a cabinet or closing a microwave, as well as handling deformable materials such as folding a tablecloth.}
    \label{fig:vis_skill}
\end{figure*}

This section presents a comprehensive qualitative evaluation comparing \MindV{} against state-of-the-art baselines. We analyze performance across three distinct regimes, including multi-stage long-horizon manipulation tasks shown in Figure~\ref{fig:vis_long}, atomic short-horizon interactions illustrated in Figure~\ref{fig:vis_short}, and generalization to complex out-of-distribution (OOD) scenarios with diverse action primitives as depicted in Figure~\ref{fig:vis_skill}.  These visualizations substantiate our quantitative findings, showcasing \MindV{}'s superiority as a physically grounded and logically coherent world simulator.

\noindent \textbf{Long-horizon Tasks.} \quad As illustrated in Figure~\ref{fig:vis_long}, baseline models exhibit a clear breakdown in long-horizon tasks. They not only violate physical common sense within a individual sub-task but also fail to maintain causal consistency across the temporal sequence. For instance, baseline attempts to interact with an object that spontaneously disappeared in a previous step highlight a profound failure in tracking latent world states. In contrast, \MindV{} maintains robust logical coherence throughout the sequence. This success stems from our hierarchical design: the Semantic Reasoning Hub (SRH) explicitly decomposes complex instructions into structured sub-tasks, while the Staged Visual Future Rollouts mechanism actively filters physically implausible transitions, effectively preventing long-term semantic drift.

\noindent \textbf{Short-horizon Tasks.} \quad The short-horizon examples in Fig.~\ref{fig:vis_short} further emphasize the physical fidelity of our approach. Even when baselines correctly identify target objects, they frequently hallucinate physically impossible interactions, such as objects levitating without contact or penetrating solid surfaces. Our GRPO-based physical alignment successfully mitigates these dynamic violations. Furthermore, \MindV{} excels at interpreting abstract commands (e.g., “clean the floor”) by autonomously inferring the correct latent action sequence (e.g., “grasping a cloth, then wiping”). This demonstrates the SRH's cognitive capacity to translate high-level intent into executable, physically grounded trajectories—a key differentiator from monolithic models that struggle with action grounding.

\noindent \textbf{Complex OOD Scenarios and Diverse Interactive Actions.} \quad Fig.~\ref{fig:vis_skill} demonstrates the robust generalization capabilities that are the hallmark of a scalable world model. Panel (a) illustrates the model's precision in visually diverse OOD environments. Whether distinguishing a specific object within severe clutter or operating in a stylized scene, \MindV{} accurately executes the intent. Crucially, the background remains strictly static despite high-frequency textures. This stability is achieved because the Behavioral Semantic Bridge (BSB) decouples semantic intent from visual appearance, providing structural, domain-invariant guidance to the diffusion process. Panel (b) highlights \MindV{}'s ability to model complex physical state transitions. The model effortlessly simulates interactions with articulated objects (e.g., cabinets) and deformable materials (e.g., cloth). Executing these actions requires deep kinematic and affordance understanding, proving that integrating the SRH's cognitive reasoning with the MVG's simulation engine creates a highly capable embodied world model.

\begin{table*}[t!]
\centering
\caption{\textbf{Detailed architecture and data flow of the Motor Video Generator (MVG).} The table shows the transformation of tensor shapes from the input video and BSB guidance through each major component. Notations: B=Batch Size, C=Channels, T=Temporal Length, H=Height, W=Width, L=Sequence Length, D=Embedding Dimension.}
\label{tab:arch_details}
\resizebox{\textwidth}{!}{%
\begin{tabular}{l l l l p{6cm}}
\toprule
\textbf{Component} & \textbf{Module} & \textbf{Input Shape} & \textbf{Output Shape} & \textbf{Key Hyperparameters} \\
\midrule
\multirow{2}{*}{3D VAE} & Encoder & [B, 3, T, H, W] & [B, 16, T/4, H/8, W/8] & Latent Channels: 16 \\
& Decoder & [B, 16, T/4, H/8, W/8] & [B, 3, T, H, W] & Symmetrical to Encoder \\
\midrule
\multirow{3}{*}{\makecell[l]{Guidance \\ Embedding}} & \makecell[l]{Guidance \\ Tensor} & BSB & [B, 128, T/4, H/8, W/8] & Encodes BSB masks \& trajectories into a dense tensor. \\
\cmidrule(l){2-5}
& Spatial Conv & [B, 128, T/4, H/8, W/8] & [B, 480, T/4, H/8, W/8] & Kernel: 3x3, Stride: 1 \\
\cmidrule(l){2-5}
& Temporal Conv & [B, 480, T/4, H/8, W/8] & [B, 1920, T/4, H/8, W/8] & Kernel: 3 (1D), Stride: 1 \\
\midrule
\multirow{4}{*}{\makecell[l]{DiT Backbone \\ (30 Blocks)}} & Patch Embedding & [B, 16, T/4, H/8, W/8] & [L, D] & Patch Size: 2x2x2, Hidden Dim (D): 1920 \\
\cmidrule(l){2-5}
& \makecell[l]{Positional \\ Encoding} & [L, D] & [L, D] & Type: 3D Sinusoidal \\
\cmidrule(l){2-5}
& \makecell[l]{Timestep \\ Embedding} & Scalar $t$ & [1, 512] & - \\
\cmidrule(l){2-5}
& \makecell[l]{Transformer \\ Blocks} & [L, D] & [L, D] & Layers: 30, Attention Heads: 30, Head Dim: 64 \\
\midrule
\textbf{Scheduler} & DDIM & Scalar $t$ & Noise Schedule & Timesteps: 50, Schedule: Linear \\
\bottomrule
\end{tabular}%
}
\end{table*}

\section{Network Architecture Details}
\label{sec:supp_arch}

This section details the architectural design and data flow of the Motor Video Generator (MVG). Built upon the CogVideoX-5B~\cite{cogvideox} foundation, the MVG employs a 3D Variational Autoencoder (VAE) and a Diffusion Transformer (DiT)~\cite{DiT} backbone. Our primary modification is the conditioning mechanism, which injects explicit spatiotemporal guidance from the Behavioral Semantic Bridge (BSB) to achieve precise kinematic control. The key module specifications and tensor transformations are summarized in Table~\ref{tab:arch_details} and elaborated below.

\noindent \textbf{Latent Space Definition.} \quad The MVG operates within a compressed latent space parameterized by a pre-trained 3D-VAE~\cite{vae}. As detailed in Table~\ref{tab:arch_details}, the VAE encoder performs spatiotemporal compression on an input video of shape $[B, 3, T, H, W]$, producing a compact latent representation of shape $[B, 16, T/4,\allowbreak H/8, W/8]$. The entire forward and reverse diffusion processes are executed within this latent space. Subsequently, the VAE decoder maps the final denoised latent back to the pixel space to render the simulated video.

\noindent \textbf{Diffusion Transformer Backbone.} \quad The core of the MVG simulation engine is a 30-block DiT with a hidden dimension of 1920. The latent video is first partitioned into non-overlapping $2 \times 2 \times 2$ spatiotemporal patches, which are linearly embedded into a sequence of tokens. This token sequence is augmented with 3D sinusoidal positional encoding and diffusion timestep embeddings prior to being processed by the transformer blocks.

\noindent \textbf{BSB Conditioning Mechanism.} \quad To enforce action adherence, the conditioning mechanism integrates BSB guidance into the DiT backbone via a Guidance Embedding module. The structured semantic information from the BSB is first rasterized into a dense Spatiotemporal Guidance Tensor, which is then processed by a sequence of spatiotemporal convolutions, specifically a $3 \times 3$ spatial convolution followed by a 1D temporal convolution, yielding a 1920-dimensional feature representation $G$. Maintaining identical spatiotemporal resolution to the latent video, $G$ is injected into the DiT backbone via additive fusion exclusively within even-numbered transformer blocks (e.g., blocks 0, 2, 4, $\dots$), leaving odd-numbered blocks unconditioned. This alternating fusion strategy interleaves explicit kinematic control from the BSB with the model's internal generative priors, steering the denoising trajectory to adhere to the planned trajectory.

\section{Key GRPO Hyperparameters}
We initialize GRPO from the SFT checkpoint and train for 1,500 iterations with AdamW using a learning rate of $5 \times 10^{-5}$, $\beta_1=0.9$, $\beta_2=0.999$, and weight decay $1 \times 10^{-4}$. For each prompt, the policy samples a group of $G=8$ candidate videos to compute group-relative advantages, avoiding an additional value network while preserving stable credit assignment. We adopt a clipping threshold of $\epsilon=0.2$ in the GRPO objective and set the KL regularization coefficient to $\beta=0.01$, which constrains overly large denoising-policy updates and keeps the updated policy close to the SFT reference model. The reward combines aesthetic and physical terms with $w_a=1.0$ and $w_p=0.2$, normalizing the two reward scales while prioritizing visual fidelity and retaining sufficient pressure toward physically coherent dynamics. For the PFC reward, we set the softmax temperature to $\tau=0.1$ so that optimization emphasizes the least plausible temporal windows without collapsing the reward to a single frame segment. Unless otherwise specified, GRPO uses 50 DDIM denoising steps and 37-frame subtask videos at $480 \times 640$ resolution, matching the SFT and evaluation settings. 

\section{User Study Protocol}
\label{sec:supp_user_study}

To complement automatic metrics, we conduct a user study to evaluate the perceptual quality and task-level correctness of long-horizon manipulation videos. The study includes 30 volunteer participants with backgrounds in computer vision, robotics, or related engineering fields. All participants were recruited on a voluntary basis, and no monetary compensation or other payment was provided. Each participant is shown a randomized subset of long-horizon tasks from the 108-sample evaluation set. For each task, we present the original instruction, the initial scene observation, and anonymized videos generated by all compared methods. Method names are hidden, and the display order is randomly shuffled for every question to avoid positional and brand bias.

Participants are asked to judge each video according to three criteria: (1) whether the video follows the language instruction and completes the intended sub-tasks in the correct order; (2) whether the object interactions are physically plausible, including contact, collision, object permanence, and motion continuity; and (3) whether the video is visually coherent, with limited artifacts, flickering, or identity drift. For each task, participants select the best overall video under these criteria. We aggregate the selections across all valid responses and report the resulting percentage as the User Preference score in Table~\ref{tab:t2}. To ensure response reliability, all videos are shown at the same resolution and playback speed, participants are allowed to replay clips before making a decision, and incomplete responses are excluded from aggregation.

\end{document}